\documentclass{article}

\usepackage{subcaption}
\usepackage{graphicx}
\usepackage{booktabs}
\usepackage{multirow}
\usepackage{amsmath}

\usepackage{makecell}
\usepackage[table]{xcolor}
\usepackage{adjustbox}
\usepackage{hyperref}
\usepackage{titletoc}

\usepackage[main, final]{neurips_2026}

\usepackage[utf8]{inputenc} 
\usepackage[T1]{fontenc}    
\usepackage{hyperref}       
\usepackage{url}            
\usepackage{booktabs}       
\usepackage{amsfonts}       
\usepackage{nicefrac}       
\usepackage{microtype}      
\usepackage{xcolor}         
\usepackage{alltt}

\definecolor{yesgreen}{HTML}{C6EFCE}
\definecolor{nored}{HTML}{FFC7CE}
\definecolor{partialyellow}{HTML}{FFEB9C}

\newcommand{\Y}{\cellcolor{yesgreen}\rule{0pt}{2.2ex}\phantom{Yes}}
\newcommand{\N}{\cellcolor{nored}\rule{0pt}{2.2ex}\phantom{Yes}}
\newcommand{\Part}{\cellcolor{partialyellow}\rule{0pt}{2.2ex}\phantom{Yes}}

\newcommand{\rot}[1]{\rotatebox[origin=c]{90}{\makecell{#1}}}

\newcommand{\appendixtoc}{
    \clearpage
    \phantomsection
    \addcontentsline{toc}{section}{Appendices}
     \begin{center}
        \rule{\linewidth}{1.6pt}\\[0.3em]
        \vspace*{1em}
        {\LARGE \textbf{Appendix}} \\[0.5em]
        \rule{\linewidth}{0.6pt}
    \end{center}
    \startcontents[sections]
    \printcontents[sections]{}{1}{}
    \begin{center}
      \rule{\linewidth}{0.4pt} 
    \end{center}
}

\title{When Can Digital Personas Reliably Approximate Human Survey Findings?}
\author{%
  Mumin~Jia\\
  Department of Mathematics and Statistics\\
  York University\\
  Toronto, Ontario M3J 1P3 \\
  \texttt{amyjia@yorku.ca} 
  \And
  Yilin~Chen\\
  Department of Mathematics and Statistics\\
  York University\\
  Toronto, Ontario M3J 1P3 \\
  \texttt{cathy929@yorku.ca} 
  \And
  Divya~Sharma\\
  Department of Mathematics and Statistics\\
  York University\\
  Toronto, Ontario M3J 1P3 \\
  Department of Biostatistics\\
  University Health Network\\
  Toronto, ON M5G 2C4 \\
  \texttt{divya03@yorku.ca} 
  \And Jairo~Diaz-Rodriguez\\
  Department of Mathematics and Statistics\\
  York University\\
  Toronto, Ontario M3J 1P3 \\
  \texttt{jdiazrod@yorku.ca}
}

\begin{document}

\maketitle

\begin{abstract}
  Digital personas powered by Large Language Models (LLMs) are increasingly proposed as substitutes for human survey respondents, yet it remains unclear when they can reliably approximate human survey findings. We answer this question using the LISS panel, constructing personas from respondents’ background variables and pre-2023 survey histories, then testing them against the same respondents’ held-out post-cutoff answers. Across four persona architectures, three LLMs, and two prediction tasks, we assess performance at the question, respondent, distributional, equity, and clustering levels. Digital personas improve alignment with human response distributions, especially in domains tied to stable attributes and values, but remain limited for individual prediction and fail to recover multivariate respondent structure. Retrieval-augmented architectures provide the clearest gains, but performance depends more on human response structure than on model choice: personas perform best for low-variability questions and common respondent patterns, and worst for subjective, heterogeneous, or rare responses. Our results provide practical guidance on when digital personas could be appropriate for survey research and when human validation remains necessary.
\end{abstract}

\section{Introduction}

Large language models are increasingly being used to simulate human survey respondents \citep{argyle2023out,bisbee2024synthetic,dominguez-olmedo2024questioning}. These ``digital personas'' promise faster, cheaper, and more flexible alternatives to conventional survey data collection, with potential uses in pre-testing survey instruments, exploring subgroup patterns, simulating hard-to-reach populations, and conducting early-stage social measurement before fielding studies with real participants. Yet their scientific value depends on more than whether their answers appear plausible: a synthetic respondent may reproduce population averages while failing to match the individuals it is meant to represent \citep{kaiser2025simulating,toubia2025twin}, select majority responses while compressing minority or rare answer patterns, or preserve marginal distributions while distorting subgroup differences, respondent heterogeneity, and multivariate structure among survey items \citep{bisbee2024synthetic,wang2025large,li2025promise}. These failures matter because survey research is often used not only to estimate aggregate opinion, but also to study variation across people, groups, domains, and latent population structure.

This paper asks: \textit{when can digital personas reliably approximate human survey findings?} We study this question in a ground-truth setting using the Longitudinal Internet Studies for the Social Sciences (LISS) panel, a probability-based longitudinal survey of Dutch households. For each respondent, we construct digital personas from information observed before a temporal cutoff and evaluate them against held-out answers from the same human respondent after that cutoff. This design tests whether persona-based simulations preserve real human responses, rather than merely generating demographically plausible or population-like answers.

\begin{figure}[!t]
    \centering
    \includegraphics[width=0.9\textwidth]{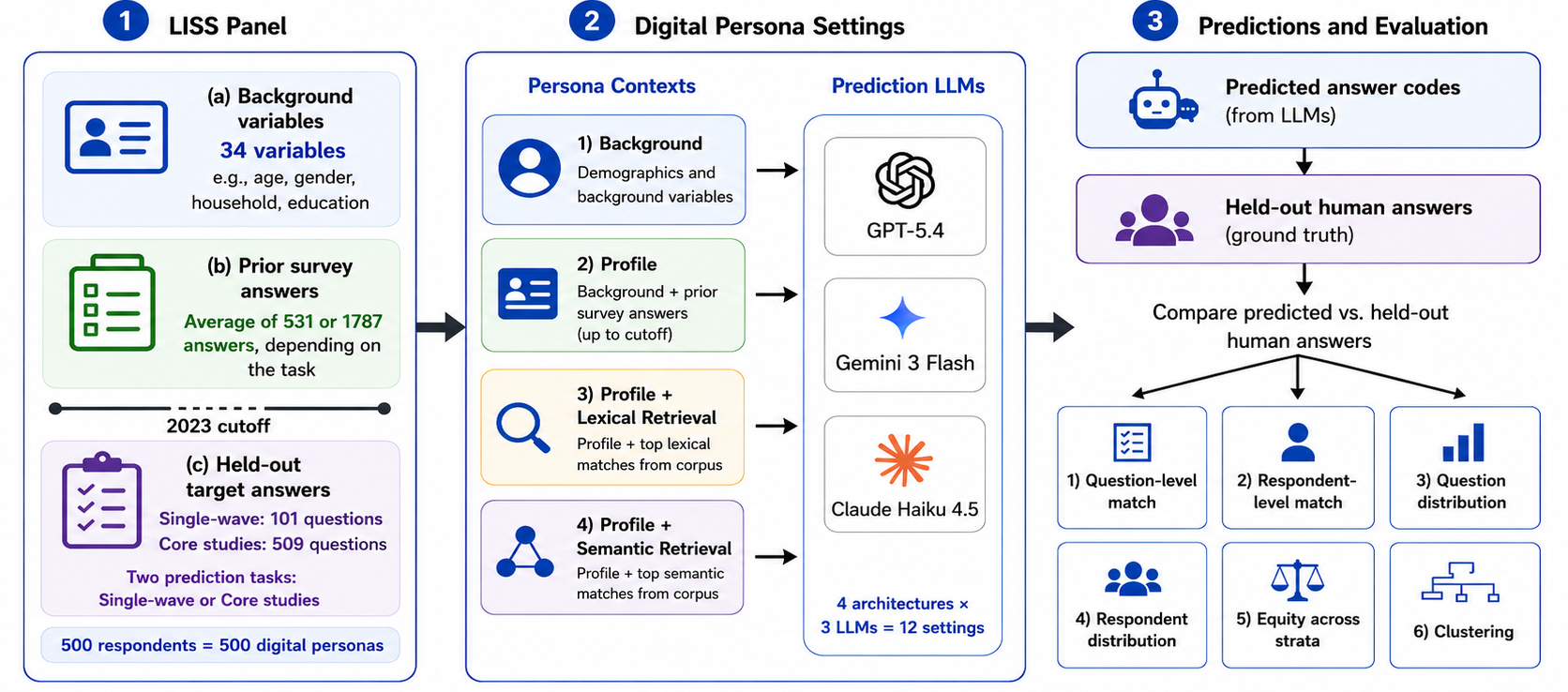}
    \caption{
    \small Overview of our digital persona evaluation framework.}
    
    \label{fig:overview}
\end{figure}

Figure~\ref{fig:overview} provides an overview of our digital persona evaluation framework. Starting from the LISS panel, we use each respondent's background variables and pre-2023 survey histories to construct a digital persona, then ask that persona to predict the respondent's held-out post-cutoff survey answers. The framework varies both the information given to the persona and the model used for prediction: personas may receive only background variables, a structured profile, or a profile augmented with lexically or semantically retrieved prior answers, and predictions are generated with multiple LLM backbones. The resulting answer codes are then compared with the same respondents' true held-out answers. Rather than treating reliability as a single accuracy score, the framework evaluates whether digital personas preserve human survey findings across six complementary dimensions: question-level match, respondent-level match, question-level distributions, respondent-level response profiles, equity across demographic strata, and respondent clustering. This design allows us to ask not only whether digital personas can predict individual answers, but also whether they preserve the aggregate and structural patterns that survey researchers use for scientific inference. Overall, our study brings together data, modeling, and evaluation dimensions that prior work has typically studied separately. Table~\ref{tab:related_work_comparison} shows that this combination has not been examined in a single empirical design.

\textbf{Summary of findings.}
We provide an exploratory analysis of each reliability dimension in
Section~\ref{sec:results}, followed by a confirmatory and explanatory modeling
analysis in Section~\ref{sec:modelling}. Together, these analyses provide practical guidance on when digital personas are likely to
be useful in survey research. First, digital personas align most closely with
human response distributions in domains tied to stable attributes, values, and
background information, such as family and household, politics and values, and
religion and ethnicity. They perform worse in domains that depend on lived
experience, self-assessment, or situational detail, such as social integration
and leisure and personality. Second, retrieval-augmented persona contexts
provide the clearest architectural benefit, while differences across the modern
LLMs we evaluate are comparatively small. Third, performance appears broadly
stable across gender, age group, and household stage, suggesting no large
systematic demographic disparities. Fourth, reliability varies more strongly across the response space: personas are most accurate for low-variability questions and common respondent patterns, and least accurate for high-variability questions and rare respondent patterns. Question structure further shapes performance, with binary and small-answer-space questions easier to approximate than subjective questions or items likely to elicit diverse responses. Overall, digital personas are most useful for survey design and
early-stage instrument testing, especially for identifying questions likely to
yield stable aggregate patterns versus those requiring validation with human
data.

\paragraph{Contributions.}
This work makes four contributions. First, we introduce a ground-truth evaluation design for digital personas in survey research using the LISS panel. Second, we propose a multi-dimensional reliability framework that evaluates personas at the question, respondent, distributional, equity, and clustering levels, thereby separating aggregate approximation from individual substitution. Third, we systematically compare four persona construction strategies and three LLM prediction models across two cross-domain prediction tasks. Fourth, we provide empirical evidence on the conditions under which digital personas appear more or less reliable, offering practical guidance for survey design and validation.

\section{Position Relative to Prior Work}

\begin{table}[t]
\centering
\scriptsize
\setlength{\tabcolsep}{3pt}
\renewcommand{\arraystretch}{1.15}
\caption{\small Comparison of our study with related work. Green indicates that the study substantially includes the feature; yellow indicates a related but narrower or incomplete version; red indicates that the feature is absent or not central.}
\label{tab:related_work_comparison}
\begin{adjustbox}{max width=\textwidth}
\begin{tabular}{lccccccccccccccc}
\toprule
\textbf{Study} &
\rot{Fixed-choice\\answers} &
\rot{Rep. stratified\\population sample} &
\rot{Same-person\\holdout} &
\rot{Temporal\\split} &
\rot{Real longitudinal\\panel} &
\rot{Prior survey\\history} &
\rot{Persona\\variants} &
\rot{Retrieval\\memory} &
\rot{Multi-LLM\\comparison} &
\rot{No-context\\baseline} &
\rot{Individual\\fidelity} &
\rot{Aggregate\\distribution} &
\rot{Equity /\\subgroups} &
\rot{Joint\\structure} &
\rot{Error\\drivers} \\
\midrule

\textbf{Our study} &
\Y & \Y & \Y & \Y & \Y & \Y & \Y & \Y & \Y & \Y & \Y & \Y & \Y & \Y & \Y \\

\citet{argyle2023out} &
\Part & \Part & \N & \N & \N & \Part & \Part & \N & \N & \Part & \N & \Y & \Part & \N & \Part \\

\citet{bisbee2024synthetic} &
\Y & \Y & \N & \N & \N & \Part & \Part & \N & \N & \Y & \N & \Y & \Part & \N & \Part \\

\citet{dominguez-olmedo2024questioning} &
\Y & \Y & \N & \N & \N & \N & \N & \N & \Y & \Y & \N & \Y & \Part & \N & \Y \\

\citet{hwang-etal-2023-aligning} &
\Y & \Part & \Part & \Part & \Part & \Y & \Part & \Part & \N & \Y & \Y & \N & \Part & \N & \Part \\

\citet{park2024generative} &
\Part & \Part & \Y & \Part & \Part & \Y & \Part & \Part & \Part & \Y & \Y & \Part & \Part & \N & \Part \\

\citet{chen-etal-2025-personatwin} &
\Part & \N & \Part & \N & \N & \Y & \Y & \Part & \Part & \Y & \Y & \Part & \Y & \N & \Part \\

\citet{jiang2025know} &
\Part & \N & \N & \Y & \N & \Part & \Y & \Y & \Y & \Y & \Part & \N & \N & \N & \Part \\

\citet{li-etal-2025-far} &
\Part & \N & \N & \Part & \N & \Part & \Y & \Part & \Y & \Y & \Part & \N & \N & \N & \Part \\

\citet{toubia2025twin} &
\Y & \Part & \Y & \Y & \Y & \Y & \N & \N & \N & \N & \Y & \Part & \Part & \N & \N \\

\citet{kaiser2025simulating} &
\Y & \Y & \Part & \N & \N & \N & \Part & \N & \Part & \Y & \Y & \Y & \Y & \N & \Part \\

\citet{cao-etal-2025-specializing} &
\Y & \Y & \N & \Part & \N & \N & \N & \N & \Y & \Y & \N & \Y & \Y & \N & \Part \\

\citet{wang2025large} &
\Part & \Y & \N & \N & \N & \N & \Part & \N & \Y & \Y & \N & \Part & \Y & \N & \Y \\

\citet{li2025promise} &
\Part & \Part & \N & \N & \N & \N & \Y & \N & \Part & \Part & \N & \Part & \Y & \N & \Y \\

\citet{abdulhai2026consistently} &
\N & \N & \N & \N & \N & \N & \Y & \Part & \Part & \Y & \Part & \N & \N & \N & \Part \\

\citet{taday2026assessing} &
\Y & \Y & \N & \N & \N & \N & \Part & \N & \Y & \Y & \N & \Y & \Y & \N & \Y \\

\citet{li2025hugagent} &
\N & \N & \Part & \Part & \N & \Part & \Y & \Part & \Y & \Y & \Y & \N & \N & \N & \Part \\

\citet{park2023generative} &
\N & \N & \N & \N & \N & \N & \Y & \Y & \N & \Y & \Part & \N & \N & \N & \Part \\

\citet{moon2024anthology} &
\Part & \Part & \N & \N & \N & \N & \Y & \N & \Part & \Y & \N & \Part & \Y & \N & \Part \\

\citet{hu2024persona} &
\Part & \N & \N & \N & \N & \N & \Y & \N & \Part & \Y & \Part & \N & \Part & \N & \Y \\

\citet{kim2023ai} &
\Y & \Y & \N & \Part & \Part & \N & \N & \N & \Part & \Y & \N & \Y & \Part & \N & \Part \\

\citet{suh2025subpop} &
\Y & \Y & \N & \Part & \N & \N & \N & \N & \Part & \Y & \N & \Y & \Y & \N & \Part \\

\citet{kolluri2025socsci210} &
\Part & \N & \N & \Part & \N & \Part & \N & \N & \Part & \Y & \Part & \Part & \Y & \N & \Y \\

\citet{hu2025simbench} &
\Part & \N & \N & \Part & \N & \N & \Part & \N & \Y & \Y & \Part & \Part & \Y & \N & \Y \\

\citet{rupprecht2025ggss} &
\Y & \Y & \N & \N & \N & \N & \Y & \N & \Part & \Y & \N & \Y & \Part & \N & \Part \\

\citet{du2025twinvoice} &
\N & \N & \Part & \Part & \N & \Part & \Y & \Part & \Y & \Y & \Y & \N & \N & \N & \Part \\

\citet{wu2026humanlm} &
\Part & \N & \Part & \Part & \Part & \Y & \Y & \Part & \Y & \Y & \Y & \Part & \Part & \N & \Part \\

\citet{chen2026omnibehavior} &
\Part & \N & \Part & \Y & \Part & \Y & \Y & \Part & \Y & \Y & \Y & \Part & \Part & \N & \Y \\

\citet{hullman2026validating} &
\N & \N & \N & \N & \N & \N & \N & \N & \N & \N & \N & \N & \Part & \N & \Y \\

\bottomrule
\end{tabular}
\end{adjustbox}
\end{table}

In Table~\ref{tab:related_work_comparison}, we position our study against prior work on synthetic survey respondents, digital personas, and persona reliability. The comparison emphasizes the combination of features needed for a strict test of digital-persona fidelity: objectively scored outcomes, a representative longitudinal panel, same-person temporal holdout prediction, respondent-specific prior survey histories, and evaluation beyond aggregate fit.

The main pattern is that prior work covers important but partial areas of this design space. Appendix~\ref{app:criteria} defines the comparison criteria, and Appendix~\ref{app:related-work-positioning} provides the full related-work discussion and study-by-study comparison. Several of the closest studies are recent preprints that have not yet appeared in peer-reviewed venues, but the overall pattern is consistent: no single study combines the full set of data, persona-construction, and evaluation elements that we examine here.

\section{Experiments}\label{sec:experiments}

Our experiments study digital personas as conditional substitutes for real human participants in held-out survey prediction. Rather than asking whether LLMs can produce plausible survey answers in isolation, we construct tasks in which  digital personas must predict the actual responses of real respondents from the LISS panel. We vary the available respondent history, the form of persona representation, the retrieval strategy, and the prediction model. This design provides a controlled setting for examining when digital personas have enough respondent-specific evidence to approximate real human answers, and when they do not.

\subsection{Setup}\label{sec:setup}

\paragraph{Dataset.}
We use the Longitudinal Internet Studies for the Social Sciences (LISS) panel, a probability-based online panel of Dutch households administered by Centerdata \citep{mulder2024liss}. The LISS panel is well suited to our setting because it repeatedly surveys the same individuals across multiple domains and waves, and also provides background variables describing respondents' sociodemographic characteristics. This structure allows us to construct a held-out prediction task: earlier responses and background information are used to build a digital persona, while later responses from the same respondent serve as ground truth.

\paragraph{Studies.}
The panel contains two broad types of questionnaires. \emph{Core studies} are repeated longitudinal modules covering relatively stable domains such as personality, health, family, work, religion, politics, and social values. \emph{Single-wave studies} are one-off surveys that often focus on topical or domain-specific questions. This distinction allows us to test generalization in both directions: from stable core histories to topical single-wave targets, and from broader single-wave histories to structured core-study targets.

\paragraph{Respondents.}
Using 2023 as the temporal cutoff, we retain respondents with a non-empty background file, non-empty prior-answer history, and at least one evaluable target answer. This yields 6,276 respondents with valid records. Each respondent is represented by an anonymized numeric identifier and has 34 sociodemographic background variables, including gender, age, household composition, urbanicity, income bracket, and related attributes.

\paragraph{Questions and responses.}
We restrict the analysis to closed-ended responses with finite answer spaces, including ordinal, true/false, nominal, and binary questions. We keep only standalone questions that can be answered from the question text alone, excluding items that require external context or refer to other survey items. The question metadata contain 2,923 unique standalone entries and specify the question text, variable name, response type, and allowed answer categories.

\paragraph{Prior and target answers.}
For each respondent, we split survey answers using 2023 as the temporal cutoff: pre-cutoff answers are used as prior evidence for persona construction, and held-out answers at or after the cutoff are used for evaluation. We define two prediction tasks:

\begin{itemize}
    \item \textbf{\emph{Single-wave prediction:}} prior evidence consists of the respondent's most recent pre-cutoff core-study answers, and targets are single-wave answers from 2023--2024.
    \item \textbf{\emph{Core prediction:}} prior evidence consists of the respondent's full pre-cutoff single-wave answer history, and targets are core-study answers from 2023.
\end{itemize}

This design evaluates each digital persona against unseen responses from the same human respondent.

\subsection{Respondent Sampling}

For each task, we sample 500 eligible respondents from the LISS panel using stratified proportional sampling. Respondents are eligible if they have non-empty background variables, prior-answer history for the relevant input scope, and at least one evaluable target answer. Strata are defined by gender, age group, and household stage; strata with fewer than five eligible respondents are merged with the nearest compatible stratum. Within each stratum, we prioritize respondents with greater answer coverage using
\[
    s_i = z(n^{\mathrm{prior}}_i) + z(n^{\mathrm{target}}_i),
\]
where $n^{\mathrm{prior}}_i$ and $n^{\mathrm{target}}_i$ denote the number of available prior and target answers for respondent $i$. The resulting samples preserve demographic variation while maintaining sufficient answer coverage; their composition is reported in Table~\ref{tab:sample_demographics} in Appendix. Detailed design is available in Appendix~\ref{app:sampling_design}. Appendix Figure~\ref{fig:question_coverage} summarizes respondent-level answer coverage in the prior and target partitions used for evaluation after sampling.

\subsection{Digital Personas}

We define a digital persona as the prompt context used by an LLM to answer held-out survey questions for a specific respondent. For each prediction call, the LLM receives a block of target questions and a respondent-specific context, and returns predicted answer codes constrained to the admissible response options. We evaluate three prediction LLMs, GPT 5.4, Gemini 3 Flash, and Claude Haiku 4.5, crossed with four persona architectures, yielding 12 digital persona settings. The architecture determines what respondent information is included in the prompt context: background variables only, a structured profile, a profile with lexical retrieval, or a profile with semantic retrieval. We describe the architectures below, with full details provided in Appendix~\ref{app:experiments_details} and our GitHub repository.

\paragraph{Background.}
The background persona context contains only the respondent's 34 background variables and no prior survey answers. This is the minimal respondent-specific prompt condition and measures how much the prediction LLM can infer from sociodemographic information alone.

\paragraph{Profile.}
The profile persona context contains the respondent's background variables and a structured natural-language summary of their prior answers. The profile is generated once using GPT-5.4. The raw prior-answer rows are not included at prediction time. This condition tests whether a compact summary of survey history preserves useful respondent-specific information.

\paragraph{Profile + lexical retrieval.}
This persona context augments the structured profile with retrieved prior answers. For each prediction sub-batch, we append the $K$ prior answered rows most lexically related to the current target questions. This condition tests whether surface-level similarity between prior and target survey items provides useful respondent-specific evidence.

\paragraph{Profile + semantic retrieval.}
This persona context uses the same structure as lexical retrieval, but replaces token overlap with embedding-based retrieval. Prior answered rows and target questions are embedded offline. For each prediction sub-batch, each prior row is scored by summing its cosine similarities to all target-question embeddings in the sub-batch. The top-$K$ prior rows are appended to the prompt. This condition tests whether semantic similarity retrieves useful respondent memories when related survey items are phrased differently.

\paragraph{\emph{Baseline.}}
As a baseline, we remove respondent-specific information entirely. The prediction LLM receives only the target question text, variable name, representation type, admissible answer categories, and answer constraints. Each target question is queried 500 times, matching the number of sampled respondents, so that the generated answer distribution can be compared with the empirical human distribution without using respondent information. This baseline separates gains from persona conditioning from performance driven by general LLM priors or majority-response tendencies. Appendix~\ref{app:baselines} clarifies the baseline design choices.

\begin{figure}[t]
    \centering
    \includegraphics[width=0.9\textwidth]{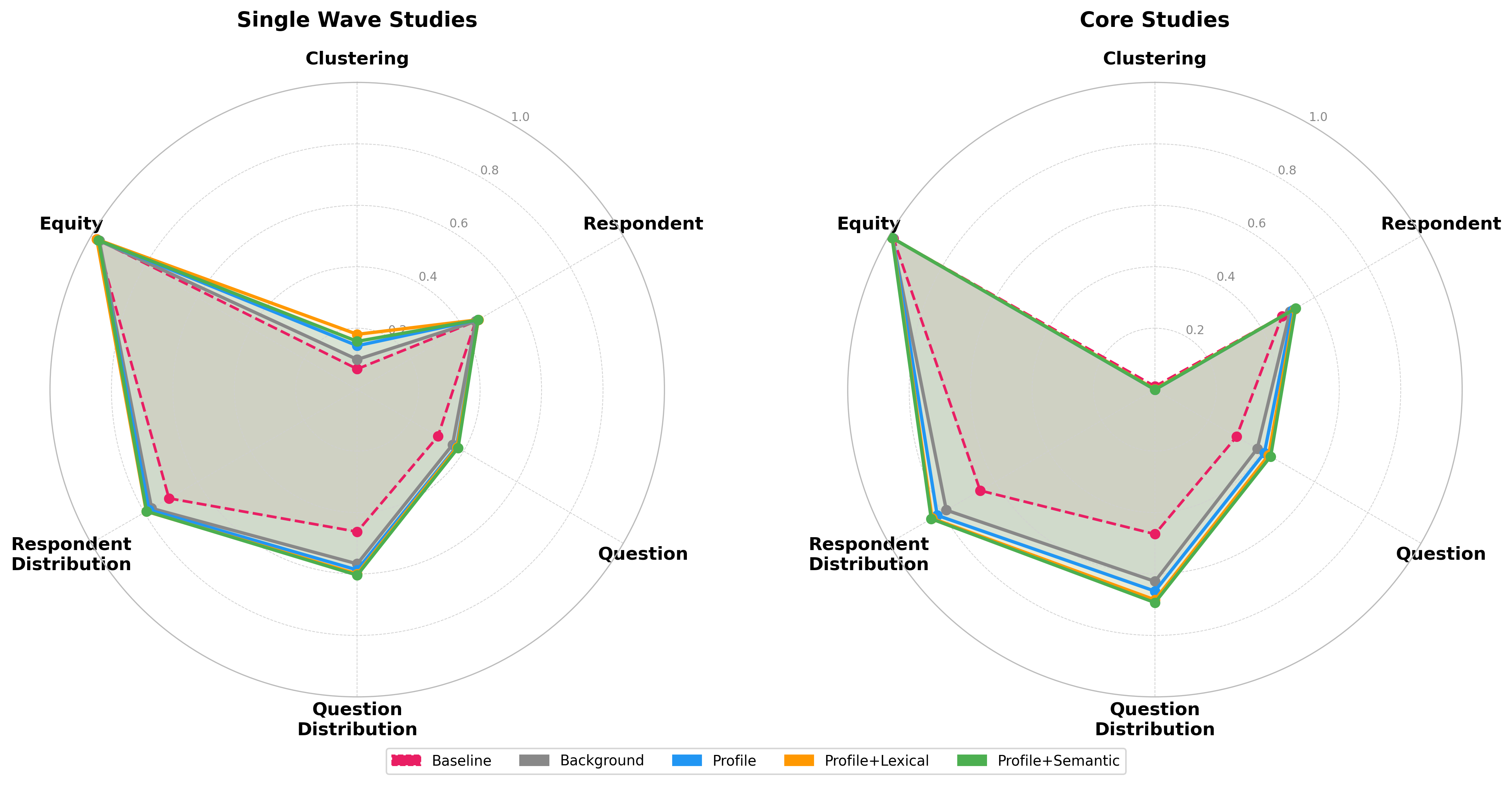}
    \caption{\small 
    Aggregate reliability of digital persona settings across evaluation dimensions. Radar plots summarize performance for the single-wave and core-study prediction tasks across question-level match, respondent-level match, question-level distributional alignment, respondent-level distributional alignment, equity, and clustering. Higher values indicate better performance.
    }
    \label{fig:radar_summary}
\end{figure}

\section{Evaluation Framework}

\begin{table}[t]
\centering
\scriptsize
\caption{\small Fidelity of digital personas on core predictions across five persona conditions (1 baseline + 4 personas) and three LLMs. Metrics measure Clustering ARI $\uparrow$, Respondent exact match rate $\uparrow$, Question weighted F1 score $\uparrow$, Question Distribution JSD $\downarrow$, Respondent Distribution MMD $\downarrow$, and equity mean absolute deviation of DPI $\downarrow$. Core studies involve richer question entries and respondents, yielding higher persona-fidelity headroom compared to single-wave predictions. Baseline is GPT-5.4 only. Bold denotes values statistically tied with the best-performing setting under overlap of the reported uncertainty intervals.}
\label{tab:cs_dimensions}
\begin{tabular}{llccccc}
\toprule
\multicolumn{2}{l}{} & \textbf{Baseline} & \textbf{Background} & \textbf{Profile} & \textbf{Profile\texttt{+}Lexical} & \textbf{Profile\texttt{+}Semantic} \\
\midrule
\multirow{3}{*}{\textbf{Clustering ARI} ($\uparrow$)}
& \textbf{GPT}    & 0.011 ($\pm$0.011) & 0.000 ($\pm$0.002) & 0.000 ($\pm$0.004) & 0.000 ($\pm$0.037) & 0.000 ($\pm$0.020) \\
& \textbf{Claude} & ---                 & 0.009 ($\pm$0.005) & 0.011 ($\pm$0.041) & \textbf{0.133 ($\pm$0.068)} & 0.000 ($\pm$0.060) \\
& \textbf{Gemini} & ---                 & 0.000 ($\pm$0.005) & 0.000 ($\pm$0.006) & 0.000 ($\pm$0.007) & 0.000 ($\pm$0.002) \\
\midrule

\multirow{3}{*}{\textbf{Resp. Match rate} ($\uparrow$)}
& \textbf{GPT}    & 0.478 ($\pm$0.002) & 0.507 ($\pm$0.002) & 0.521 ($\pm$0.002) & 0.529 ($\pm$0.002) & 0.531 ($\pm$0.002) \\
& \textbf{Claude} & ---                 & 0.486 ($\pm$0.002) & 0.497 ($\pm$0.002) & 0.510 ($\pm$0.002) & 0.510 ($\pm$0.002) \\
& \textbf{Gemini} & ---                 & 0.515 ($\pm$0.002) & 0.528 ($\pm$0.002) & \textbf{0.536 ($\pm$0.002)} & \textbf{0.534 ($\pm$0.002)} \\
\midrule

\multirow{3}{*}{\textbf{Question F1 Score} ($\uparrow$)}
& \textbf{GPT}    & 0.306 ($\pm$0.003) & 0.392 ($\pm$0.004) & 0.419 ($\pm$0.003) & 0.434 ($\pm$0.003) & \textbf{0.436 ($\pm$0.004)} \\
& \textbf{Claude} & ---                 & 0.367 ($\pm$0.003) & 0.399 ($\pm$0.003) & 0.412 ($\pm$0.004) & 0.415 ($\pm$0.003) \\
& \textbf{Gemini} & ---                 & 0.402 ($\pm$0.004) & 0.423 ($\pm$0.004) & \textbf{0.444 ($\pm$0.004)} & \textbf{0.444 ($\pm$0.004)} \\
\midrule

\multirow{3}{*}{\textbf{Ques.~Dist.~JSD} ($\downarrow$)}
& \textbf{GPT}    & 0.530 ($\pm$0.004) & 0.375 ($\pm$0.005) & 0.338 ($\pm$0.005) & 0.315 ($\pm$0.005) & \textbf{0.306 ($\pm$0.005)} \\
& \textbf{Claude} & ---                 & 0.409 ($\pm$0.004) & 0.370 ($\pm$0.004) & 0.345 ($\pm$0.004) & 0.338 ($\pm$0.004) \\
& \textbf{Gemini} & ---                 & 0.365 ($\pm$0.005) & 0.339 ($\pm$0.005) & \textbf{0.305 ($\pm$0.005)} & \textbf{0.301 ($\pm$0.005)} \\
\midrule

\multirow{3}{*}{\textbf{Resp.~Dist.~MMD} ($\downarrow$)}
& \textbf{GPT}    & 0.343 ($\pm$0.002) & 0.215 ($\pm$0.002) & 0.181 ($\pm$0.002) & 0.161 ($\pm$0.002) & \textbf{0.155 ($\pm$0.002)} \\
& \textbf{Claude} & ---                 & 0.241 ($\pm$0.002) & 0.202 ($\pm$0.002) & 0.176 ($\pm$0.002) & 0.179 ($\pm$0.002) \\
& \textbf{Gemini} & ---                 & 0.204 ($\pm$0.002) & 0.184 ($\pm$0.002) & 0.165 ($\pm$0.002) & \textbf{0.157 ($\pm$0.002)} \\
\midrule

\multirow{3}{*}{\textbf{Equity DPI} ($\downarrow$)}
& \textbf{GPT}    & 0.017 ($\pm$0.003) & 0.015 ($\pm$0.002) & 0.013 ($\pm$0.002) & \textbf{0.012 ($\pm$0.002)} & \textbf{0.011 ($\pm$0.002)} \\
& \textbf{Claude} & ---                 & 0.022 ($\pm$0.002) & 0.016 ($\pm$0.002) & 0.014 ($\pm$0.002) & \textbf{0.011 ($\pm$0.002)} \\
& \textbf{Gemini} & ---                 & \textbf{0.007 ($\pm$0.002)} & \textbf{0.009 ($\pm$0.003)} & \textbf{0.009 ($\pm$0.002)} & \textbf{0.009 ($\pm$0.002)} \\
\bottomrule
\end{tabular}
\end{table}

We evaluate digital persona reliability from six complementary dimensions: exact match at the question and respondent levels, distributional alignment at the question and respondent levels, equity across demographic strata, and preservation of respondent clusters. Together, these analyses distinguish individual-level prediction from aggregate survey approximation and test whether persona errors are evenly distributed or concentrated in particular groups or response structures.

\paragraph{Question dimension.}
We evaluate exact-response prediction at the question level. For each target question, we compare the answers generated by digital personas with the corresponding human answers across respondents. Performance is measured using weighted F1-score. This angle helps identify which survey questions are reliably approximated and which item characteristics are associated with higher or lower performance.

\paragraph{Respondent dimension.}
We evaluate prediction accuracy at the respondent level by measuring how well each digital persona reproduces the corresponding human respondent's answers across target questions. For each respondent, we compute exact match rate over the questions they answered. This angle helps identify which respondents are easier or harder to approximate and allows us to study whether fidelity varies with respondent characteristics.

\paragraph{Question-Distribution dimension. }
We evaluate whether digital personas preserve population-level response distributions. For each target question, we compare digital persona answer distributions with empirical human answer distributions for the same question. We measure this distance using Jensen--Shannon divergence (JSD) \eqref{eq:JSD} \citep{JSD_61115}. Lower JSD indicates closer alignment between synthetic and human response distributions. This captures whether personas preserve aggregate survey findings even when individual-level exact matches are imperfect.

\paragraph{Respondent-distribution dimension. }
We evaluate whether digital personas preserve each respondent's overall response profile across questions. Because each respondent is represented by a multivariate distribution over answers across many survey items, we use maximum mean discrepancy (MMD) to compare the digital persona response distribution with the human response distribution  \eqref{eq:MMD} \citep{MMD_gretton12a}. This angle captures whether the persona preserves the respondent's overall pattern of answer usage, such as agreement, disagreement, extremity, or category preference.

\paragraph{Equity dimension.}
We evaluate whether persona reliability varies across the demographic strata used for respondent sampling. Specifically, we compare performance across gender, age group, and household-stage categories using the mean absolute deviation of the Demographic Parity Index (DPI) across strata \eqref{eq:EB}. This analysis measures whether prediction errors are distributed evenly across groups or concentrated within particular strata. It is important because aggregate performance may appear stable even when reliability differs systematically across groups.

\paragraph{Clustering dimension.}
We evaluate whether digital persona responses preserve higher-level respondent structure. We cluster respondents using their observed human answers and separately cluster digital personas using their generated answers, then compare the resulting partitions using Adjusted Rand Index (ARI) \eqref{eq:ARI} \citep{ARI_Hubert1985}. Higher ARI indicates stronger agreement between human-based and persona-based respondent clusters. This angle helps test whether personas preserve multivariate population structure relevant for segmentation, typology discovery, and subgroup analysis.

\section{Results}\label{sec:results}

We generate held-out responses for all digital persona settings and the no-context baseline in both 2023 prediction tasks: core-study prediction and single-wave prediction. We now discuss the main findings from these predictions.

\begin{figure}[t]
    \centering

    \begin{minipage}[t]{0.7\textwidth}
        \centering
        \includegraphics[width=\textwidth]{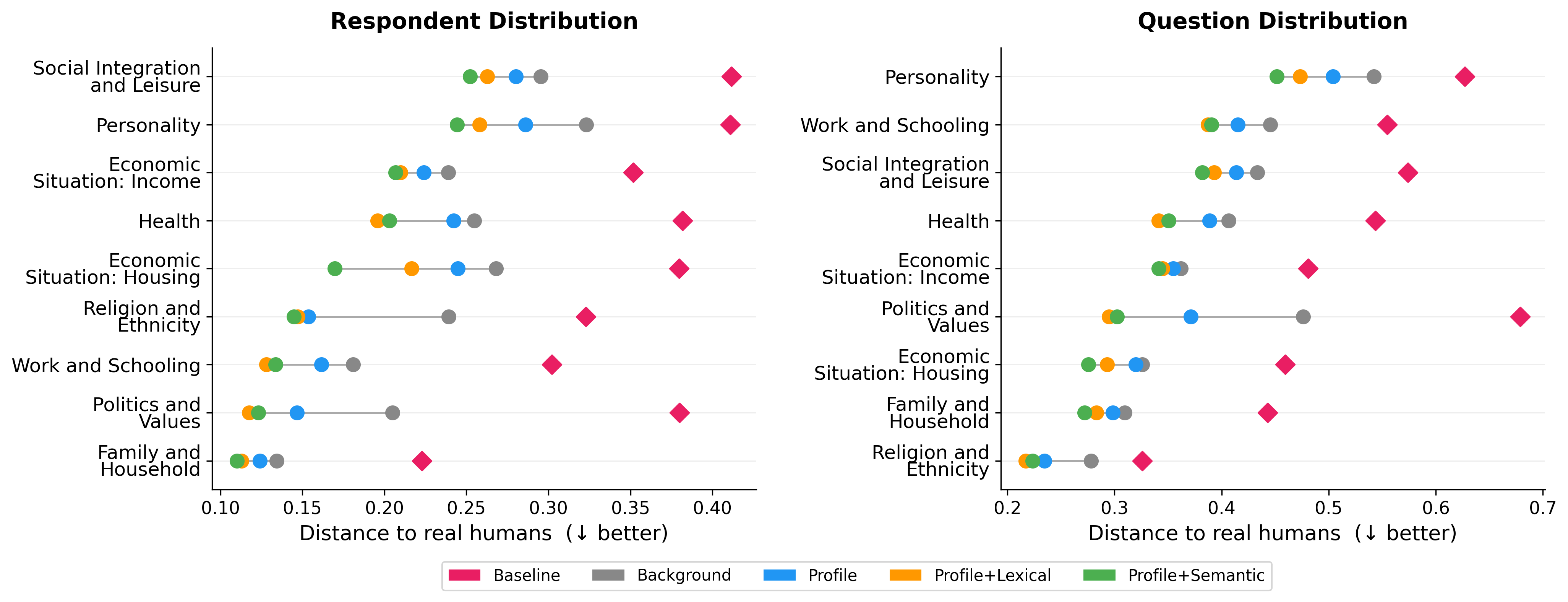}
    \caption{
    \small Distributional distance to human responses by study domain. Lower values indicate closer alignment with the empirical human distribution. The left panel compares respondent-level response distributions, while the right panel compares question-level response distributions.
    }
    \label{fig:distribution_lollipop}
    \end{minipage}
    \hfill
    \begin{minipage}[t]{0.28\textwidth}
        \centering
        \includegraphics[width=\textwidth]{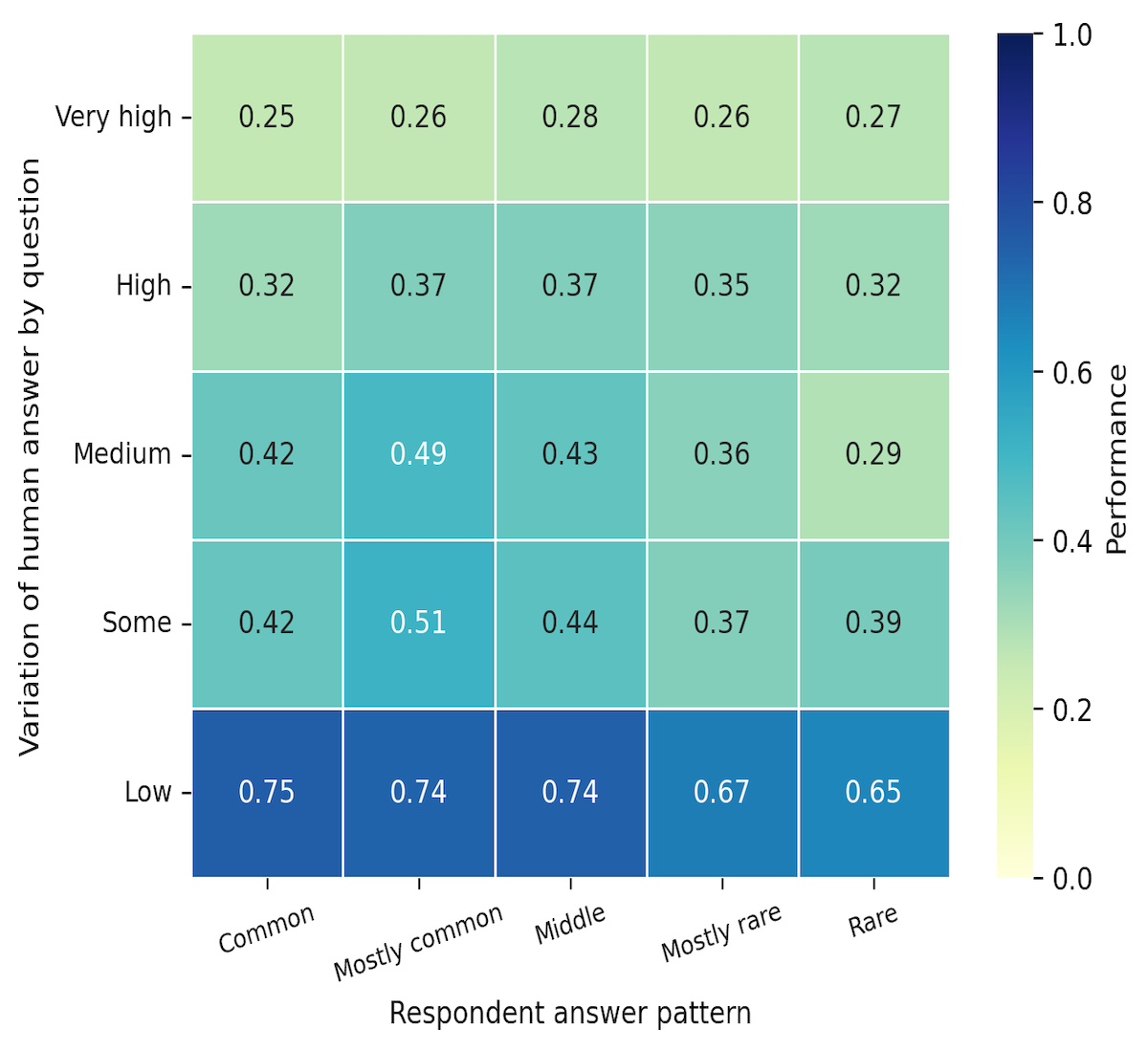}
        \caption{\small Digital persona performance by question-level answer variability and respondent answer pattern.} 
        \label{fig:heatmap}
    \end{minipage}

\end{figure}

\subsection{Overview of Persona Reliability}

Figure~\ref{fig:radar_summary} summarizes performance across the six evaluation dimensions for both prediction tasks. All metrics are scaled from 0 to 1, with higher values indicating better performance. Overall, the results show a consistent pattern. Equity scores are stable across settings, suggesting no large disparities across the demographic strata used in our sample. Clustering performance is weak across all settings, indicating that persona-generated responses do not reliably recover the respondent structure induced by human answers. Question-level and respondent-level exact-match performance improve over the baseline, but remain limited, suggesting that digital personas are not reliable substitutes for individual respondents. In contrast, the clearest advantage appears in distributional alignment: persona-based settings more closely approximate human response distributions and consistently outperform the no-context baseline. Table~\ref{tab:cs_dimensions} reports the corresponding core-study results across all architectures and LLMs, including bootstrapped confidence intervals, while the corresponding single-wave results are provided in Table~\ref{tab:sw_dimensions} in the appendix. The results show that retrieval-augmented digital personas are consistently among the best-performing settings across nearly all metrics, with differences supported by the reported confidence intervals. Across prediction LLMs, gains are small compared with the gains from persona architecture and retrieval. We therefore interpret model choice as secondary. 

\subsection{Distributional Alignment}

Figure~\ref{fig:distribution_lollipop} compares the distance between human response distributions and the distributions produced by each setting for the core-study prediction task. Across both respondent-level and question-level views, the baseline is consistently farther from humans than the digital personas, indicating that baseline predictions may exploit majority-answer patterns but do not reproduce the empirical distribution of human responses. 
Digital personas substantially reduce distributional distance across most domains, with retrieval variants consistently closest to the human distributions. 

\begin{figure}[t!]
    \centering
    \includegraphics[width=0.95\textwidth]{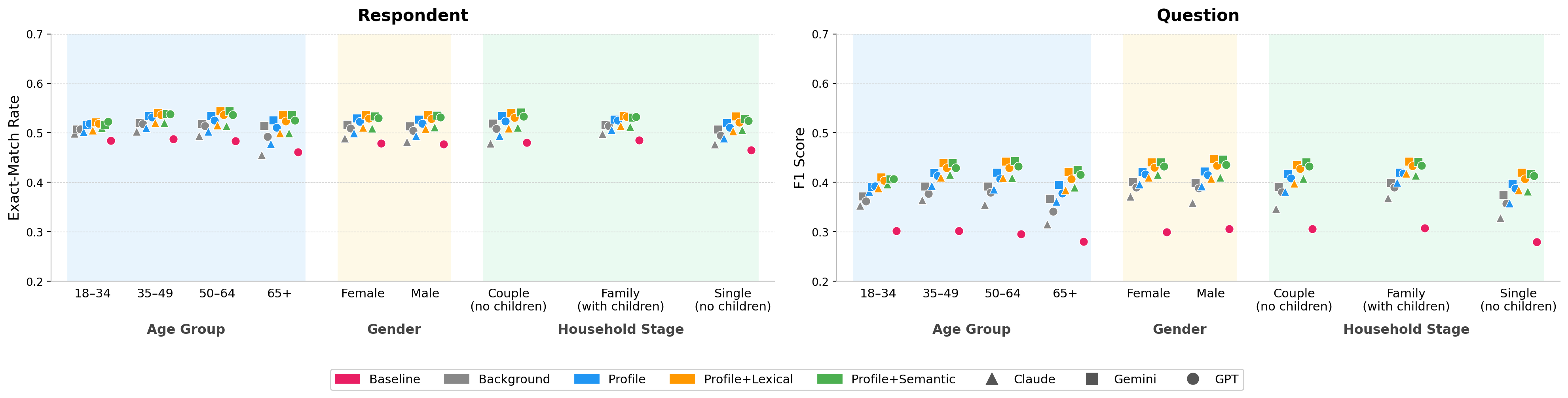}
        \caption{\small Exact-match performance across demographic strata at the respondent level (left, measured by exact match rate) and question level (right, measured by weighted F1 score).}
        \label{fig:demographic_breakdown}
        \vspace{-3mm}
\end{figure}

Digital personas align most closely with human distributions in family and household, politics and values, and religion and ethnicity. These domains may be easier to approximate because they are closely tied to information available in the persona context. Family and household responses are often reflected in background variables such as age, marital status, children, and household composition. Politics, values, religion, and ethnicity tend to be more stable over time and may be captured by prior attitude- and identity-related survey answers.
In contrast, social integration and leisure and personality show larger distributional distances. These domains depend more on individual lived experience and self-assessment, rather than broad demographic or ideological signals. As a result, they are harder for digital personas to infer unless the prior history contains directly related evidence.

\subsection{Equity and Conditional Exact-Match Performance}

Figure~\ref{fig:demographic_breakdown} reports performance across the demographic strata used in sampling. Respondent-level performance (left) and question-level performance (right) are broadly stable across gender, age group, and household stage, with no large systematic disparities. This suggests that persona performance is not driven by a single demographic group and is relatively balanced across the sampled strata. Across architectures, retrieval-augmented settings perform slightly best overall. Across LLMs, Gemini performs slightly better than GPT, while Claude performs weakest.

Figure~\ref{fig:heatmap} shows a different source of variation. Performance changes substantially when conditioned on question-level human answer variability and respondent answer commonness. Accuracy is highest for low-variability questions and common respondent patterns, and lowest for high-variability questions and rare respondent patterns. Thus, while performance appears demographically balanced, persona reliability is still uneven: errors concentrate where human responses are more heterogeneous and less common. Details on the construction of the heatmap are provided in Appendix~\ref{app:heatmap}.

\subsection{Clustering and Multivariate Respondent Structure}

Figure~\ref{fig:ari_overall} in the appendix evaluates whether digital persona responses recover the respondent clusters obtained from human answers. Across settings, clustering agreement is generally low, indicating that synthetic response matrices do not preserve the same multivariate respondent structure as the human data. This provides an important caution: even when digital personas improve question- or respondent-level distributional alignment, they may still fail to preserve higher-order dependence patterns among answers. More details and additional analysis available in Appendix~\ref{app:clustering}.

\section{What Predicts Whether a Digital Persona Matches a Human Response?}\label{sec:modelling}

\begin{figure}[t]
    \centering
    \includegraphics[width=\textwidth]{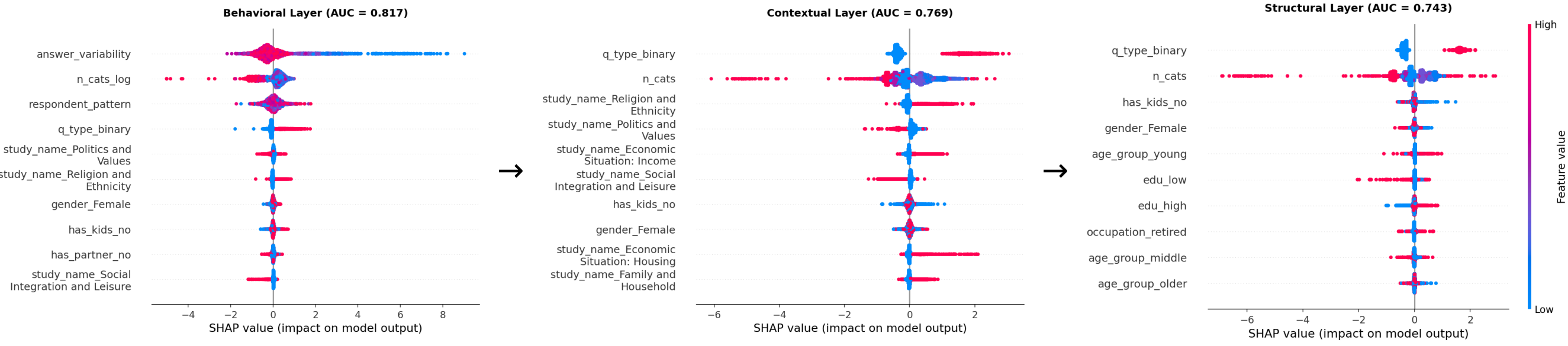}
    \caption{\small 
Top 10 predictors of digital persona accuracy from XGBoost models across three explanatory feature layers. Each point is a respondent-question prediction; the x-axis shows the SHAP value, where positive values increase the model's predicted probability of a correct persona answer and negative values decrease it. Color indicates the feature value, from low to high. The behavioral layer includes empirical response-distribution features, the contextual layer removes these distribution-derived variables, and the structural layer further focuses on observable respondent and question features.
}
    \label{fig:shap_cs}
\end{figure}

The results above suggest that digital personas are more reliable for distributional approximation than for exact individual prediction, with performance varying across study domains and regions of the response space. We use respondent-question-level modeling as a confirmatory and explanatory analysis, testing whether these patterns remain predictive of individual-level accuracy when question, study, and respondent features are considered jointly. For this analysis, we focus on the GPT-5.4 profile with lexical retrieval setting. Each observation is one predicted answer for one respondent on one target question, with a binary outcome indicating whether the digital persona exactly matches the human answer. We compare logistic regression, mixed-effects logistic regression, decision trees, random forests, and XGBoost across three feature layers: behavioral, contextual, and structural. Table~\ref{tab:sw_cs_comparison} reports the predictive performance for core-study prediction. Tree-based models perform best overall, with XGBoost achieving the highest accuracy across all three feature layers while remaining among the strongest models in AUC. We therefore use XGBoost for the feature-attribution analysis in Figure~\ref{fig:shap_cs}, which reports the top ten SHAP features from the best XGBoost models for core prediction. Features are ordered by importance, so variables appearing near the top contribute most to the model's predictions. Additional modeling details, bootstrapped confidence intervals, and corresponding results for single-wave prediction are provided in Appendix~\ref{app:modelling}.

\begin{table}[t]
\centering
\caption{\small Core study prediction performance across feature layers ($n=132{,}887$ observations).}
\label{tab:sw_cs_comparison}
\setlength{\tabcolsep}{5pt}
\begin{tabular}{l cc cc cc}
\toprule
 & \multicolumn{2}{c}{\textbf{Behavioral}} & \multicolumn{2}{c}{\textbf{Contextual}} & \multicolumn{2}{c}{\textbf{Structural}} \\
\cmidrule(lr){2-3} \cmidrule(lr){4-5} \cmidrule(lr){6-7}
\textbf{Model} & ACC & AUC & ACC & AUC & ACC & AUC \\
\midrule
Logistic Regression & 0.706 ($\pm$0.001) & 0.775 & 0.658 ($\pm$0.002) & 0.722 & 0.632 ($\pm$0.002) & 0.700 \\
MEGLM               & 0.705 ($\pm$0.001) & 0.765    & 0.653 ($\pm$0.001) & 0.719 & 0.643 ($\pm$0.001) & 0.700 \\
Decision Tree       & 0.699 ($\pm$0.001) & 0.802 & 0.681 ($\pm$0.002) & 0.738 & 0.672 ($\pm$0.002) & 0.732 \\
Random Forest       & 0.727 ($\pm$0.002) & \textbf{0.819} & 0.695 ($\pm$0.002) & 0.758 & 0.681 ($\pm$0.002) & \textbf{0.744} \\
XGBoost             & \textbf{0.757} ($\pm$0.002) & 0.817 & \textbf{0.705} ($\pm$0.002) & \textbf{0.769} & \textbf{0.685} ($\pm$0.002) & 0.743 \\
\bottomrule
\end{tabular}
\end{table}

\paragraph{Behavioral layer.}
The behavioral layer includes empirical response-structure features from Figure~\ref{fig:heatmap}, including question-level answer variability and respondent answer pattern. These features dominate the model. Answer variability is the strongest predictor of persona accuracy, followed by the number of answer categories and respondent answer pattern. This supports the result from Figure~\ref{fig:heatmap}: personas are most accurate when human answers are concentrated and respondent patterns are common, and least accurate when questions elicit heterogeneous or rare responses. Question format also matters: larger answer spaces reduce performance, while binary questions increase it.

\paragraph{Contextual layer.}
The contextual layer removes behavioral predictors but retains broader survey context, including the substantive domain and format of the target question. This approximates a prospective setting in which empirical response distributions are unavailable, but the general meaning of the question is known. In this layer, question structure remains highly predictive: larger answer spaces reduce performance, while binary questions increase it. Domain context is also important, consistent with Figure~\ref{fig:distribution_lollipop}: personas perform better in domains tied to stable attributes, values, and background information, and worse in domains that depend more on lived experience, self-assessment, or situational detail.

\paragraph{Structural layer.}
The structural layer further removes study identifiers, leaving question form and respondent attributes. Question type and number of answer categories remain the strongest predictors, while respondent characteristics have smaller effects. This qualifies the demographic results in Figure~\ref{fig:demographic_breakdown}: performance is broadly balanced across sampled strata, but demographics still explain some residual variation once stronger behavioral and contextual predictors are removed.

Overall, this analysis ties the empirical results back to the central question of the paper: when can digital personas approximate human survey findings? By identifying the conditions under which persona predictions are more or less reliable, it provides a guide for practical use and motivates the conclusion below.

\section{Conclusion} \label{sec:conclusion}

Our findings provide a practical guideline for using digital personas in survey research. Practitioners should use them most cautiously for individual-level prediction and most confidently for aggregate distributional approximation, especially when questions have small answer spaces and human responses are likely to be concentrated. Digital personas appear more reliable in domains tied to stable attributes, values, and background information, such as family and household, politics and values, and religion and ethnicity, and less reliable in domains that depend on lived experience, self-assessment, or situational detail. Their performance also appears broadly balanced across the demographic strata we study, suggesting that errors are not driven mainly by gender, age group, or household stage. Across the modern LLMs we evaluate, performance differences are relatively small, while retrieval-augmented persona contexts provide the clearest architectural benefit. Notably, the limited gains from more sophisticated persona architectures are themselves informative: they suggest that the primary bottleneck for digital persona reliability is not the richness of the retrieval mechanism or the choice of model, but the structure of the response space itself. However, reliability varies strongly with the structure of the response task: personas are less appropriate when the target requires recovering rare, heterogeneous, subjective, or highly respondent-specific answers. In practice, persona reliability should therefore be assessed against the intended inferential use and the structure of the survey task. The more variable, subjective, or idiosyncratic the expected human responses are, the more important it is to validate persona outputs against real human data. 

\textbf{Limitations. }Our evaluation is conducted on a single longitudinal panel of Dutch households, and findings may not generalize to panels from other countries, languages, or cultural contexts. The framework is designed to be applicable to other longitudinal surveys with repeated measurements of the same respondents, but cross-panel validation remains important future work. The analysis is restricted to closed-ended questions with finite answer spaces, leaving open how digital personas perform on open-ended or scale-based items. Our respondent sample is capped at 500 per task due to computational constraints, and results may differ with larger or more diverse samples. Finally, persona profiles are generated using GPT-5.4. Future work should examine whether profile quality varies meaningfully across profile-generating models.

\begin{ack}
This work was supported by the Natural Sciences and Engineering Research Council of Canada (NSERC) under grants DGECR-2022-04531 and RGPIN-2024-05548. The authors thank Jianan Lei, the anonymous
reviewers, and the session chair for their valuable feedback and insightful suggestions, which greatly
improved the quality of this work.
\end{ack}



\newpage
\bibliographystyle{plainnat}
\bibliography{NeurIPS_2026/neurips_2026}

\appendix

\appendixtoc

\newpage
\section{Data Use and License Statement}\label{app:liscense}

This study uses data from the LISS panel. In accordance with the required acknowledgement, we state: ``In this paper, we make use of data from the LISS panel (Longitudinal Internet studies for the Social Sciences) managed by the non-profit research institute Centerdata (Tilburg University, the Netherlands).''

The LISS panel data are not redistributed with this manuscript. Access to the data is governed by Centerdata's LISS Data Archive rules and user statement. Published datasets can be downloaded free of charge only for non-commercial scientific or policy-relevant purposes, subject to registration and acceptance of the LISS data user statement. Access is personal, and data users are not permitted to make copies of the data available to others.

Researchers who wish to access the data should request access directly through the LISS Data Archive at \url{https://www.lissdata.nl}. After creating an account and joining the LISS community, users can search for and download the relevant LISS panel datasets, subject to Centerdata's access conditions and user statement.

Nevertheless, all results reported in this manuscript are fully reproducible by users who obtain permission to access the relevant LISS panel data and use the accompanying replication materials provided by the authors.

All analyses, interpretations, and conclusions in this manuscript are the responsibility of the authors and do not necessarily reflect the views of Centerdata, Tilburg University, or the LISS panel.

\section{Experimental Details}
\label{app:experiments_details}

We report complete implementation details for prediction batching, prompt
construction, retrieval scoring, answer validation, and model configuration.
Code and replication materials are available at \url{https://github.com/muminjia/digital-persona-suppl}.
\subsection{Prediction Batching}
\label{app:batching}

Target questions are grouped by their originating study (\texttt{study\_id}
for single-wave targets; \texttt{project\_number} for core targets) and
partitioned into sub-batches.  For each group $s$ containing $N_g^{(s)}$
questions and a target sub-batch size $b = 20$, the number of sub-batches
and the realized sub-batch size are

\begin{equation}
  B^{(s)} = \left\lceil \frac{N_g^{(s)}}{b} \right\rceil,
  \qquad
  N_{\mathrm{batch}}^{(s)} = \left\lceil \frac{N_g^{(s)}}{B^{(s)}} \right\rceil.
  \label{eq:batching}
\end{equation}

Each API call covers exactly one sub-batch.  For the two retrieval-augmented
agents, the top-$K$ retrieval count is set dynamically to
$K = N_{\mathrm{batch}}^{(s)}$, coupling context size to the number of
questions being predicted in that call.

\subsection{Prompt Templates}
\label{app:prompts}

All four agents share the same system prompt and output schema; they differ
only in what respondent evidence appears in the user turn.

\paragraph{System prompt.}
Every API call uses the system message: \textit{``You map user profile data
to survey answers carefully and transparently.''}

\paragraph{Output schema.}
Each prediction call must return a JSON object containing a
\texttt{predictions} array, where each entry maps a \texttt{variable\_name}
to its \texttt{predicted\_answer}.  No additional keys, markdown, or prose
are permitted.

\subsubsection{Background Agent}
\label{app:prompt-baseline}

The Background agent receives only the 34-variable background file as context;
no prior survey answers are included.  The user-turn prompt is:

\begin{quote}
\begin{alltt}\small
You are simulating a specific person answering a survey.

Answer each question as this person would, using only the
respondent background data provided below.

Background data:
\{user_context\}

Instructions:
- Answer as the person.
- Use the background data to infer the most likely response.
- If the answer is uncertain, choose the response most
  consistent with the available evidence.
- Provide exactly one answer per question.

Return valid JSON only. No markdown, code fences, or extra text.

Questions:
\{questions_text\}
\end{alltt}
\end{quote}

\subsubsection{Profile Agent: Profile Generation}
\label{app:prompt-profile-gen}

Before prediction, a separate LLM call condenses the respondent's prior
answered rows into a structured persona profile.  Profile generation uses
GPT-5.4 and is executed once per respondent per input scope; the resulting
text is cached on disk and reused by all subsequent prediction calls for
that respondent, including those of the two retrieval-augmented agents.
The profile-generation prompt, where \texttt{\{history\_text\}} is the full
serialization of all prior answered rows in the selected input scope, is:

\begin{quote}
\begin{alltt}\small
You are analyzing a set of questions and answers from a single person.

Build a predictive persona profile that allows another model to
estimate how this person would respond to future survey questions
on any topic.

From the provided Q&A, extract stable behavioral patterns across:
- beliefs and worldview
- personality traits
- reasoning and decision-making style
- knowledge domains and expertise
- tolerance for uncertainty
- recurring assumptions and cognitive biases
- communication style and priorities

Do not summarize individual answers. Identify consistent rules
and tendencies that generalize across questions.

Structure your output under the following seven headings:
  1. Personality traits
  2. Reasoning style
  3. Knowledge profile
  4. Values and motivations
  5. Biases and heuristics
  6. Decision patterns
  7. Confidence patterns

Past Q&A:
\{history_text\}
\end{alltt}
\end{quote}

\noindent
The profile must contain all seven section headers and must not exceed
1{,}500 words.  If either constraint is violated, the call is retried with
the validation error appended to the prompt; generation is attempted up to
three times before the run aborts.  A cached profile is identified by user
ID, dataset source, cutoff year, input scope, and profile model; if a
matching file exists on disk, the generation call is skipped entirely.

\subsubsection{Profile Agent: Prediction Prompt}
\label{app:prompt-profile-pred}

The profile agent's prediction prompt replaces the raw prior-answer rows
with the synthesized persona profile:

\begin{quote}
\begin{alltt}\small
You are simulating a specific person answering a survey.

Answer each question as this person would, using:
  1. the behavioral profile below
  2. the respondent background data as supporting context

Background data:
\{bg_context\}

Behavioral profile:
\{structured_profile\}

Instructions:
- Answer as the person.
- Use the profile and background data to infer the most likely
  response.
- If the answer is uncertain, choose the response most consistent
  with the available evidence.
- Provide exactly one answer per question.

Return valid JSON only. No markdown, code fences, or extra text.

Questions:
\{questions_text\}
\end{alltt}
\end{quote}

\subsubsection{Retrieval-Augmented Agents: Prediction Prompt}
\label{app:prompt-topk}

Both retrieval-augmented agents (\textit{Profile\,+\,Lexical\,Top-$K$} and
\textit{Profile\,+\,Semantic\,Top-$K$}) use an identical prediction prompt;
they differ only in how the $K$ retrieved rows in
\texttt{\{related\_rows\_text\}} are selected
(Sections~\ref{app:lexical}--\ref{app:semantic}):

\begin{quote}
\begin{alltt}\small
You are simulating a specific person answering a survey.

Answer each question as this person would, using:
  1. the behavioral profile below
  2. the retrieved prior answers as behavioral evidence
  3. the respondent background data as supporting context

Background data:
\{bg\_context\}

Behavioral profile:
\{structured\_profile\}

Retrieved prior answers (most relevant to this batch):
\{related\_rows\_text\}

Instructions:
- Answer as the person.
- Use the profile, background data, and retrieved answers to
  infer the most likely response.
- If the answer is uncertain, choose the response most consistent
  with the available evidence.
- Provide exactly one answer per question.

Return valid JSON only. No markdown, code fences, or extra text.

Questions:
\{questions\_text\}
\end{alltt}
\end{quote}

\subsection{Lexical Retrieval}
\label{app:lexical}

The \textit{Profile\,+\,Lexical\,Top-$K$} agent ranks prior answered rows
by surface-form overlap with the prediction sub-batch.

\paragraph{Tokenization.}
For each question — whether a target question or a candidate prior row —
the \texttt{variable\_label} and \texttt{categories} fields are
concatenated, lowercased, and tokenized on non-alphanumeric boundaries.
Tokens of length $\leq 1$ and tokens belonging to a 40-token stopword list
are discarded; the remaining tokens form a set.

\paragraph{Pairwise scoring.}
For each candidate prior row, we count token overlap with the target question
separately over the question label (weighted $\times 3$) and the category
field (weighted $\times 1$), then add a bonus of $+2$ if the two questions
share the same representation type and $+1$ if they share the same question
type.

\paragraph{Batch-level aggregation and selection.}
Each candidate prior row is scored against the sub-batch by summing its
pairwise relevance scores across all target questions in that sub-batch.
Rows are ranked by this aggregate score in descending order, with ties
broken by original row order, and the top $K = N_{\mathrm{batch}}^{(s)}$
rows are appended to the prediction prompt.

\subsection{Semantic Retrieval}
\label{app:semantic}

The \textit{Profile\,+\,Semantic\,Top-$K$} agent replaces the lexical
scorer with cosine similarity over dense embeddings.

\paragraph{Precomputation.}
Embeddings for all prior-answer row metadata and all target-question
metadata are computed offline using OpenAI \texttt{text-embedding-3-small}.  Each entry is keyed by \texttt{variable\_name} and
stored in a sibling JSON file alongside the corresponding metadata CSV.

\paragraph{Runtime retrieval.}
At inference time, each candidate prior row is scored against the sub-batch
by summing its cosine similarity to every target question embedding in that
sub-batch.  Rows are ranked by this aggregate similarity in descending order,
and the top $K = N_{\mathrm{batch}}^{(s)}$ are appended to the prediction
prompt, following the same selection procedure as the lexical agent.

\subsection{Answer Validation and Retry}
\label{app:validation}

Every model response undergoes a four-stage validation pipeline before
results are written to disk.

\begin{enumerate}
  \item \textbf{Parsing.}  Optional Markdown code fences are stripped; the
    remaining text must parse as valid JSON.
  \item \textbf{Schema.}  The parsed object must contain a
    \texttt{predictions} array whose entries each carry
    \texttt{variable\_name} and \texttt{predicted\_answer} keys.
  \item \textbf{Coverage.}  The set of returned \texttt{variable\_name}
    values must exactly equal the expected set for the sub-batch; both
    omissions and unexpected additions are treated as failures.
  \item \textbf{Type and range.}  Each answer is normalized by declared
    response type: categorical answers must be a valid category code; numeric answers must fall within the declared range;
    open-ended answers are accepted as-is.
\end{enumerate}

\noindent
On failure, the invalid response and the associated error message are
appended to the prompt, and the call is reissued.  Prediction calls are
retried up to four times; profile-generation calls up to three times.  Any
sub-batch that exhausts all retries is recorded as a missing prediction and
counted as incorrect in accuracy computation.

\subsection{Models}
\label{app:models}

Three prediction models are evaluated: \textbf{GPT-5.4} (OpenAI),
\textbf{Gemini-3-Flash-Preview} (Google), and \textbf{Claude Haiku~4.5}
(Anthropic).  All are queried without explicit chain-of-thought reasoning.
Profile generation uses \textbf{GPT-5.4} exclusively across all agent
configurations and both prediction tasks; differences in prediction accuracy
across models therefore reflect the prediction model alone, not variation in
the persona representation each receives.  Semantic retrieval embeddings are
produced by the OpenAI \texttt{text-embedding-3-small} model.

\subsection{Clarification on Baselines}\label{app:baselines}

Because our goal is prospective survey approximation, we use baselines that do not require observing human responses to the target questions. We therefore do not include empirical majority-class, demographic-cell majority, or other target-distribution baselines as deployable baselines: these methods require access to the same held-out human answers that the digital personas are intended to approximate. Such comparisons would be oracle references rather than usable baselines in the intended setting. Our no-context LLM baseline instead measures how well the model can reproduce target response distributions without any respondent-specific information, thereby isolating the incremental value of persona conditioning under the same information constraints faced at deployment.

We also distinguish baselines from post-processing strategies applied to persona outputs. Procedures such as majority voting across repeated persona generations, ensembling predictions from multiple persona settings, or calibrating generated responses after prediction may improve downstream estimates, but they operate on the outputs of the persona system rather than providing an independent comparison method. We therefore treat such aggregation or calibration procedures as post-processing choices and leave them outside the scope of the present evaluation framework, which focuses on the reliability of the persona constructions themselves.


\section{Additional Experiments and Results}\label{app:add_exp}

\subsection{Supplementary Results for Section~\ref{sec:results}}

\subsubsection{Detailed Results of All Dimensions}\label{app:detailedresults}

Tables~\ref{tab:cs_dimensions} and~\ref{tab:sw_dimensions} report the full per-dimension
breakdown of persona fidelity for single-wave and core predictions across
five persona conditions and three LLMs. Six metrics are included: Clustering, Respondent, Question, Ques.~Dist.~JSD, Resp.~Dist.~MMD, and Equity Bias. All values are mean
$\pm$ bootstrap SE over $N{=}100$ respondent resamples, with the Baseline
condition restricted to GPT. Bold
indicates the best value per metric.

\begin{itemize}

\item \textbf{Clustering (ARI).}
$K$-means is applied independently to the ground-truth response
matrix $\mathbf{X}$ and the predicted matrix
$\hat{\mathbf{X}}$, yielding cluster assignments vectors
$\mathbf{c}$ and $\hat{\mathbf{c}}$. The Adjusted Rand Index (ARI) measures
the agreement between the two partitions:
\begin{equation}
    \mathrm{ARI}(\mathbf{c},\,\hat{\mathbf{c}}) =
\frac{\mathrm{RI} - \mathbb{E}[\mathrm{RI}]}{\max(\mathrm{RI}) - \mathbb{E}[\mathrm{RI}]},
\end{equation}\label{eq:ARI}
where RI is the Rand Index. Higher is better ($\uparrow$).

\item \textbf{Question Distribution (JSD).}
The Jensen--Shannon distance between the predicted response distribution
$\hat{P}_q$ and the ground-truth distribution $P_q$ for question $q$,
averaged over all questions:
\begin{equation}
\overline{\mathrm{JSD}} = \frac{1}{Q}\sum_{q=1}^{Q}
\sqrt{\frac{1}{2}\mathrm{KL}(P_q \| M_q) + \frac{1}{2}\mathrm{KL}(\hat{P}_q \| M_q)},
\end{equation}\label{eq:JSD}
where $M_q = \tfrac{1}{2}(P_q + \hat{P}_q)$. Lower is better ($\downarrow$).

\item \textbf{Respondent Distribution (MMD).}
The Maximum Mean Discrepancy between the true respondent embedding matrix
$\mathbf{X}$ and the predicted matrix $\hat{\mathbf{X}}$, computed via an
RBF kernel $k$:
\begin{equation}
\mathrm{MMD} = \sqrt{\frac{1}{2}\bigl(\mathbb{E}[k(\mathbf{x},\mathbf{x}')] +
\mathbb{E}[k(\hat{\mathbf{x}},\hat{\mathbf{x}}')] -
2\,\mathbb{E}[k(\mathbf{x},\hat{\mathbf{x}})]\bigr)}.
\end{equation}\label{eq:MMD}
Lower is better ($\downarrow$).

\item \textbf{Equity (Mean Absolute deviation of DPI).}
The mean absolute deviation of the Demographic Parity Index across
demographic subgroups $g \in \mathcal{G}$ (gender, age, household stage):
\begin{equation}
\mathrm{EB} = \frac{1}{|\mathcal{G}|}\sum_{g \in \mathcal{G}}
\left|\frac{\mathrm{ACC}_g}{\mathrm{ACC}_{\mathrm{all}}} - 1\right|,
\end{equation}\label{eq:EB}
where $\mathrm{ACC}_g$ is respondent-level accuracy for subgroup $g$ and
$\mathrm{ACC}_{\mathrm{all}}$ is the overall accuracy. A value of 0 indicates
perfect demographic parity. Lower is better ($\downarrow$).

\end{itemize}

Compared to the radar plots in Figure~\ref{fig:radar_summary}, which provide
an aggregate visual overview of each condition's fidelity profile, the tables
supply exact numerical values and bootstrap uncertainty estimates that the
radar cannot convey. Both representations converge on the same conclusion:
persona-rich conditions (\textit{Profile}, \textit{Profile$+$Lexical}, and
\textit{Profile$+$Semantic}) consistently outperform
\textit{Baseline} across most fidelity dimensions.

\begin{figure}[t]
    \centering
    \includegraphics[width=\textwidth]{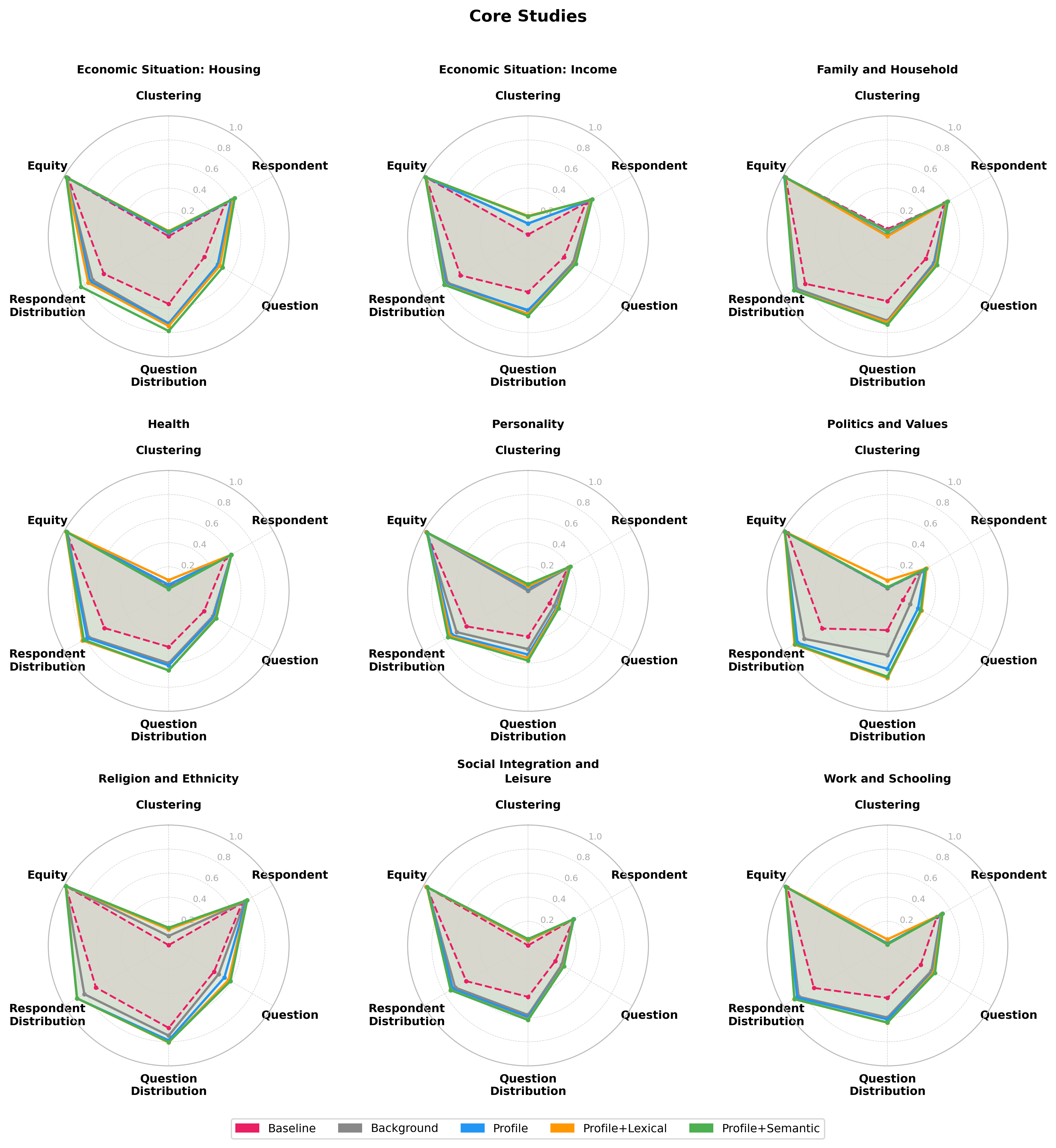}
    \caption{
    \small Per-study radar plots for core-study prediction across nine study domains. Each radar summarizes performance across the six evaluation dimensions for all persona settings. Distributional alignment gains from persona conditioning are visible across most domains, but are largest for Religion and Ethnicity and Economic Situation studies. Clustering performance remains near zero across all domains and all settings, confirming that the failure to recover respondent cluster structure is not domain-specific but a consistent limitation of persona-generated responses.
    }
    \label{fig:cs_radar_summary}
\end{figure}

\begin{figure}[t]
    \centering
    \includegraphics[width=\textwidth]{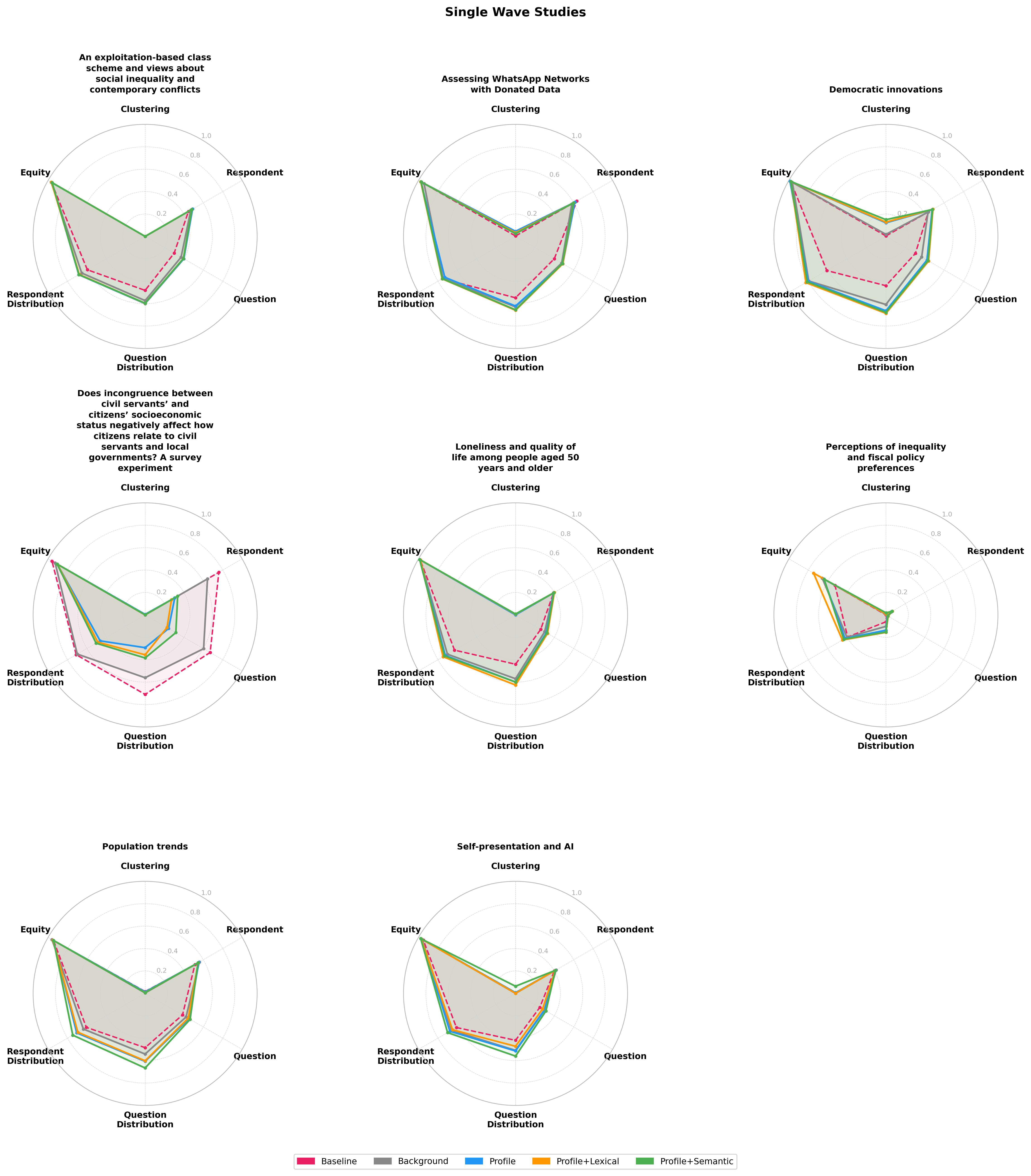}
    \caption{\small 
    Per-study radar plots for single-wave prediction across eight study domains. The pattern of results mirrors core studies: distributional alignment shows the clearest improvement over the no-context baseline across domains, while exact-match and equity metrics remain broadly stable. Clustering agreement is consistently weak regardless of study domain or persona architecture, reinforcing the finding that persona-generated responses do not reliably recover the respondent grouping structure present in human data. Studies with more concentrated answer distributions, such as Democratic Innovations, show comparatively stronger question-level match performance.
    }
    \label{fig:sw_radar_summary}
\end{figure}

We further examine whether the aggregate reliability patterns vary by study domain. Figures~\ref{fig:cs_radar_summary} and~\ref{fig:sw_radar_summary} report per-study radar plots for the core-study and single-wave prediction tasks, respectively. These plots show that the main conclusions are not driven by a small number of domains. Across studies, persona conditioning most consistently improves distributional alignment, while exact-match performance changes more modestly and clustering agreement remains weak. The per-study results therefore reinforce the broader aggregated findings.

\subsubsection{Additional Details on Figure~\ref{fig:heatmap}}\label{app:heatmap}
This heatmap examines where the best-performing digital-persona setting succeeds or fails across two forms of prediction difficulty. The plotted setting is \textit{Profile + semantic top-k} with \textit{Gemini 3 Flash} for Single Wave. 

The x-axis captures the respondent's answer pattern. For each question, we first calculate how common each real human answer is. For a respondent-question pair, answer rarity is defined as

\[
\text{answer rarity} = 1 - \text{share of respondents who gave the same answer}.
\]

We then average this rarity score across all answered questions for each respondent. Respondents are divided into five rank-based bins, from \textit{Common} to \textit{Rare}. Thus, respondents on the left tend to give common human answers, while respondents on the right tend to give rarer answer patterns.

The y-axis captures variation in the real human answers for each question. For each question, we compute the entropy of the human answer distribution and normalize it by the maximum possible entropy for that question:

\[
\text{answer variability} =
\frac{\text{entropy of human answer distribution}}
{\text{maximum possible entropy}}.
\]

Questions are then divided into five rank-based bins, from \textit{Low} to \textit{Very high}. Low-variation questions are those where humans mostly agree, while very-high-variation questions are those where humans are highly split across answer options.

Each heatmap cell combines one respondent answer-pattern bin with one question-variation bin. Within each cell, performance is computed using question-level F1 scores and then averaged across questions in that cell. This avoids letting cells with more respondent-question rows dominate only because they contain more observations. The color scale ranges from 0 to 1, with darker colors indicating higher predictive performance.

The heatmap shows that performance is strongest in the lower-left region, where questions have low human-answer variation and respondents have common answer patterns. For example, the \textit{Low} variation and \textit{Common} respondent-pattern cell reaches approximately 0.75. Performance declines as questions become more variable and as respondent answer patterns become rarer, indicating that digital personas are most reliable for ordinary questions and ordinary respondents, but less reliable for heterogeneous questions or respondents with uncommon answer profiles.

\subsubsection{Clustering}\label{app:clustering}

For the clustering analysis, each respondent was treated as a multivariate answer profile rather than evaluating one question at a time. For each analysis unit, respondents' real answers and corresponding predicted answers were encoded into the same feature space. Binary and multiple-choice questions were represented as categorical answer features.
We then used k-means clustering to partition respondents into groups based on the similarity of their answer profiles. We used up to \(k = 7\) clusters based on the silhouette score. 

The same clustering logic is used across all clustering figures, but the analysis unit changes depending on the plot. For each digital-persona architecture and large language model, clustering is performed separately on that setting's predicted answers and compared to the real-human clustering using the adjusted Rand index (ARI). ARI measures whether the predicted-answer clusters group respondents in the same way as the real-human clusters.

\begin{figure}[t]
    \centering
    \includegraphics[width=\textwidth]{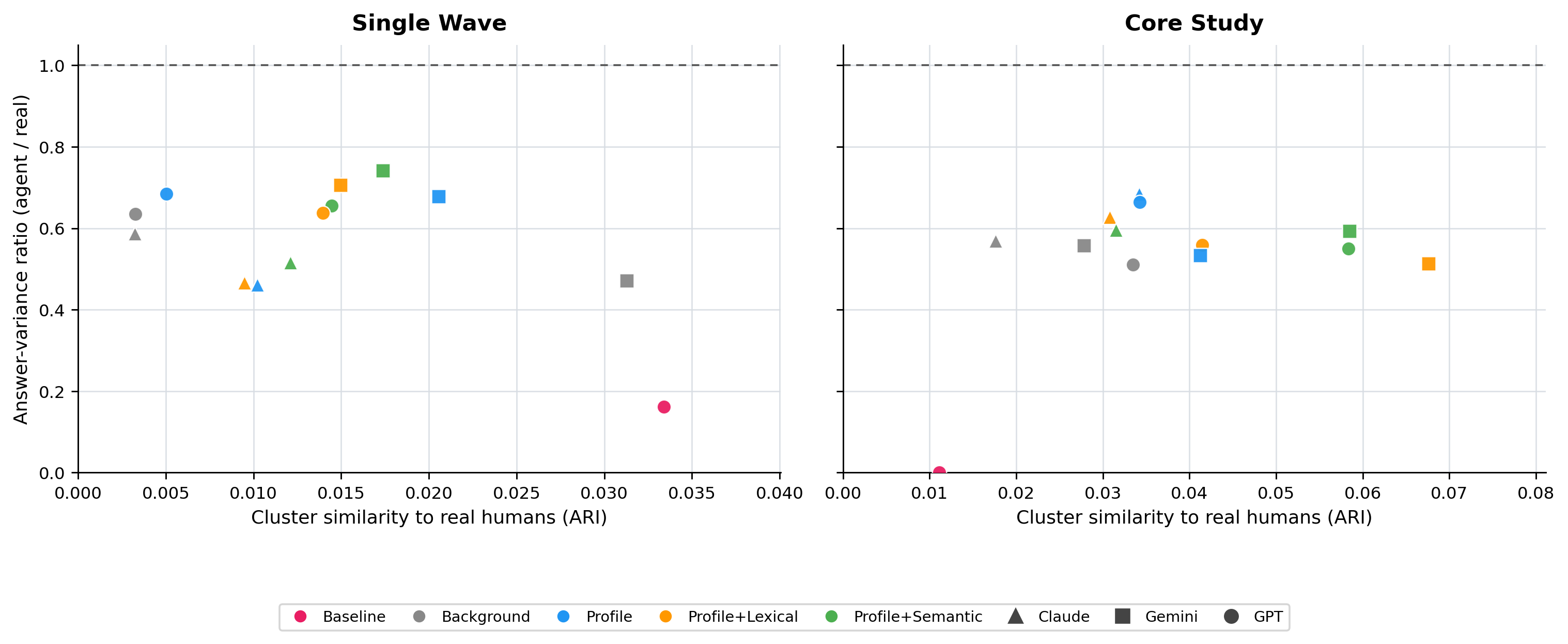}
    \caption{\small Clustering similarity and answer-variance preservation for agent-generated responses. Each point represents one digital persona configuration, with color indicating the prompting architecture and shape indicating the large language model. The x-axis reports similarity between agent and human cluster structure using ARI; higher values indicate closer recovery of human clustering. The y-axis reports the ratio of answer variance in agent responses relative to real human responses, where 1.0 indicates equal variance and lower values indicate compression of human response diversity. Panels show results separately for Single Wave and Core Study samples.}
    \label{fig:ari_overall}
\end{figure}

Figure~\ref{fig:ari_overall} summarizes this comparison at the model-setting level. The Single Wave panel has lower ARI overall than the Core Study panel, with most points below about 0.035. However, several Single Wave settings have relatively high answer-variance ratios, around 0.6 to 0.75. This suggests that some settings preserve the amount of answer variation better than they preserve the actual grouping of respondents. The Core Study panel shows somewhat higher ARI values, reaching around 0.06 to 0.07 for some settings. But the answer-variance ratios remain below 0.7, meaning that even when the clustering structure is somewhat closer to humans, the predicted answers still compress human variation. The main observation is that digital personas do not preserve clustering structure equally across topics. 

\begin{figure}[t]
    \centering
    \includegraphics[width=\textwidth]{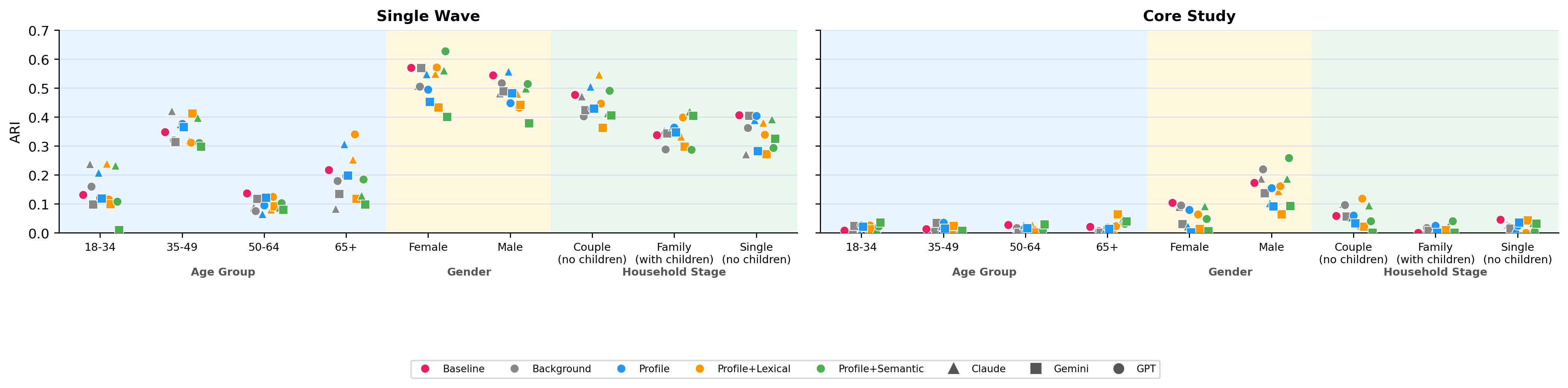}
    \caption{\small Clustering similarity by demographic subgroup. Each point reports the adjusted Rand index between real-human clusters and agent-generated clusters within a demographic subgroup. Clustering is recomputed separately for each subgroup rather than estimated once on the full sample. Colors indicate agent architecture and shapes indicate language model. Higher ARI values indicate stronger preservation of human multivariate response structure.}
    \label{fig:ari_by_demographic}
\end{figure}

For the demographic breakdown, clustering is recomputed separately within each demographic subgroup. Figure~\ref{fig:ari_by_demographic} shows that clustering preservation varies strongly by subgroup. In the Single Wave panel, gender and household-stage groups have noticeably higher ARI than many age groups. Female and Male subgroups have the highest ARI values, often around 0.4 to 0.6 depending on setting, while the 50-64 age group is much lower. In the Core Study panel, ARI values are generally much lower than in the Single Wave demographic panel. Most age-group values are close to zero, and household-stage values are also low. The clearest exception is gender, especially Male, where several settings reach higher ARI values than other Core Study subgroups. So the takeaway is that predicted answers recover demographic subgroup structure unevenly. 

\begin{figure}[t!]
    \centering
    \includegraphics[width=\textwidth]{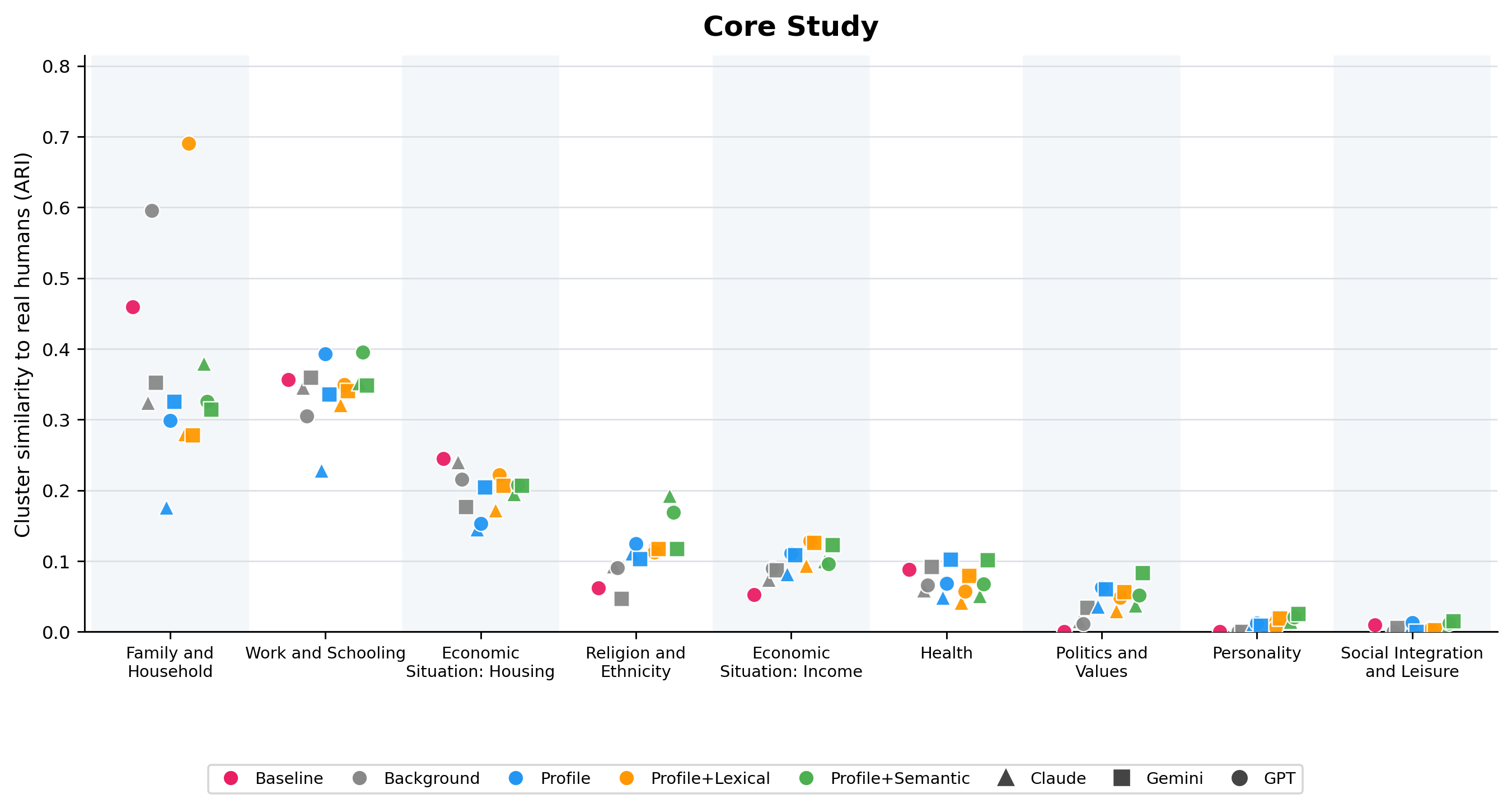}
    \caption{\small Core Study clustering similarity by topic area. Points show adjusted Rand index between real-human clusters and agent-generated clusters within each Core Study domain. Each domain is clustered separately using the respondent answer profiles available for that domain. Colors indicate agent architecture and shapes indicate language model. Higher values indicate stronger recovery of human response-pattern structure.}
    \label{fig:ari_cs}
\end{figure}

\begin{figure}[t]
    \centering
    \includegraphics[width=\textwidth]{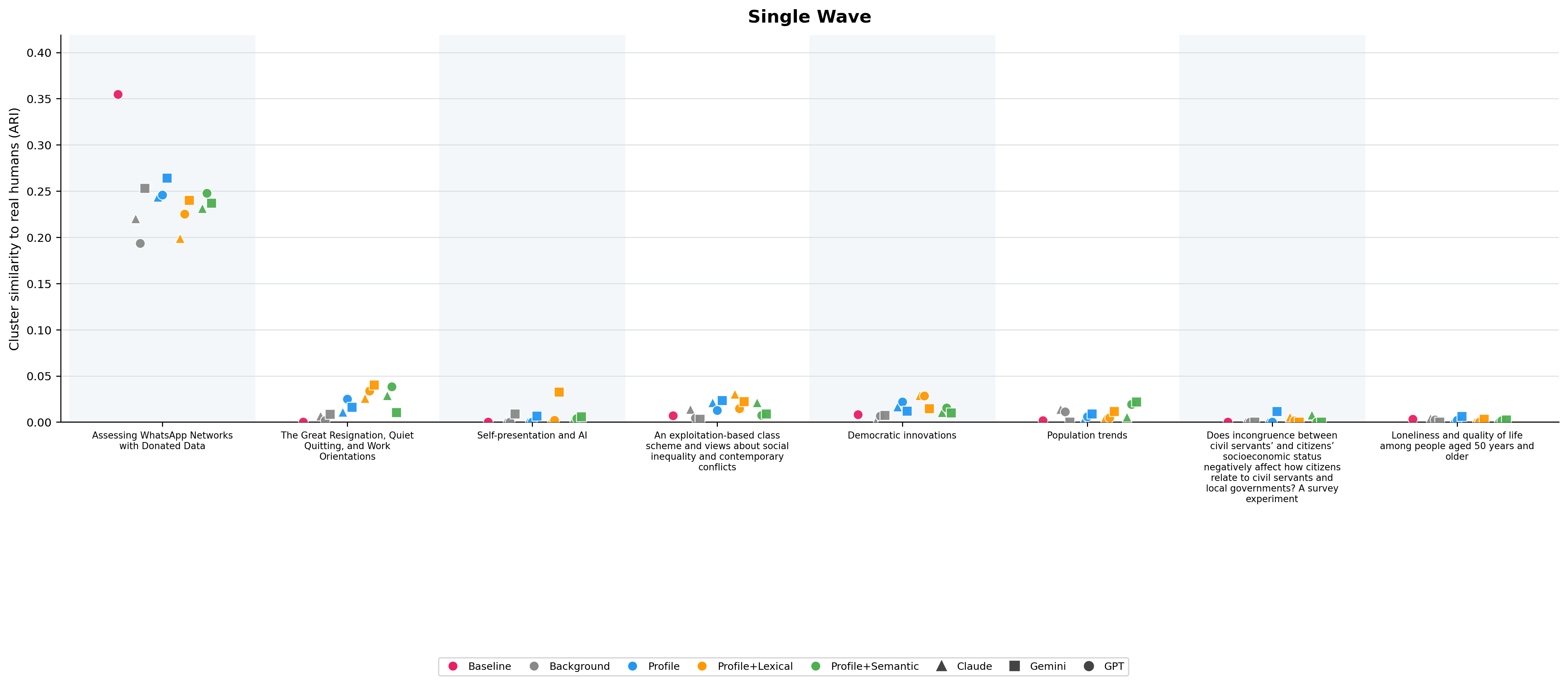}
    \caption{\small Single Wave clustering similarity by study. Points show adjusted Rand index between real-human clusters and agent-generated clusters for each Single Wave study. Clustering is computed separately within each study using respondent-level answer profiles. Colors indicate agent architecture and shapes indicate language model. Higher ARI values indicate stronger preservation of human multivariate response structure.}
    \label{fig:ari_sw}
\end{figure}

For the study-level plots, clustering is recomputed separately within each study or domain. In Figure~\ref{fig:ari_cs}, digital personas do not preserve clustering structure equally across topics. Clustering recovery is highest in Family and Household, where some settings reach around 0.6 to 0.7 ARI. Work and Schooling has moderate ARI values, roughly around 0.3 to 0.4 for several settings. Economic Situation: Housing is lower, around 0.15 to 0.25. In Figure~\ref{fig:ari_sw}, clustering preservation is highly study-specific. Assessing WhatsApp Networks with Donated Data. This study has much higher ARI than the others, with the best setting reaching around 0.35. Several model/settings in this study also cluster around 0.2 to 0.26, which is still far above the rest of the Single Wave studies. Most other Single Wave studies have ARI values near zero or only slightly above zero. The takeaway is that Single Wave clustering preservation is highly study-specific.


\begin{table}[ht]
\centering
\caption{\small Single-wave prediction performance across three feature layers ($n=30{,}725$).}
\label{tab:cs_sw_comparison}
\setlength{\tabcolsep}{5pt}
\begin{tabular}{l cc cc cc}
\toprule
 & \multicolumn{2}{c}{\textbf{Behavioral}} & \multicolumn{2}{c}{\textbf{Contextual}} & \multicolumn{2}{c}{\textbf{Structural}} \\
\cmidrule(lr){2-3} \cmidrule(lr){4-5} \cmidrule(lr){6-7}
\textbf{Model} & ACC & AUC & ACC & AUC & ACC & AUC \\
\midrule
Logistic Regression & 0.671 ($\pm$0.003) & 0.744 & 0.618 ($\pm$0.003) & 0.692 & 0.612 ($\pm$0.003) & 0.611 \\
MEGLM               & 0.670 ($\pm$0.003) & 0.744 & 0.617 ($\pm$0.003) & 0.692 & 0.612 ($\pm$0.003) & 0.607 \\
Decision Tree       & 0.685 ($\pm$0.003) & 0.757 & 0.637 ($\pm$0.003) & 0.699 & 0.625 ($\pm$0.003) & 0.667 \\
Random Forest       & 0.715 ($\pm$0.003) & 0.773 & 0.679 ($\pm$0.003) & 0.737 & 0.647 ($\pm$0.003) & 0.686 \\
XGBoost             & \textbf{0.761} ($\pm$0.003) & \textbf{0.762} & \textbf{0.693} ($\pm$0.003) & \textbf{0.725} & \textbf{0.652} ($\pm$0.003) & \textbf{0.670} \\
\bottomrule
\end{tabular}
\end{table}

\begin{figure}[t]
    \centering
    \includegraphics[width=\textwidth]{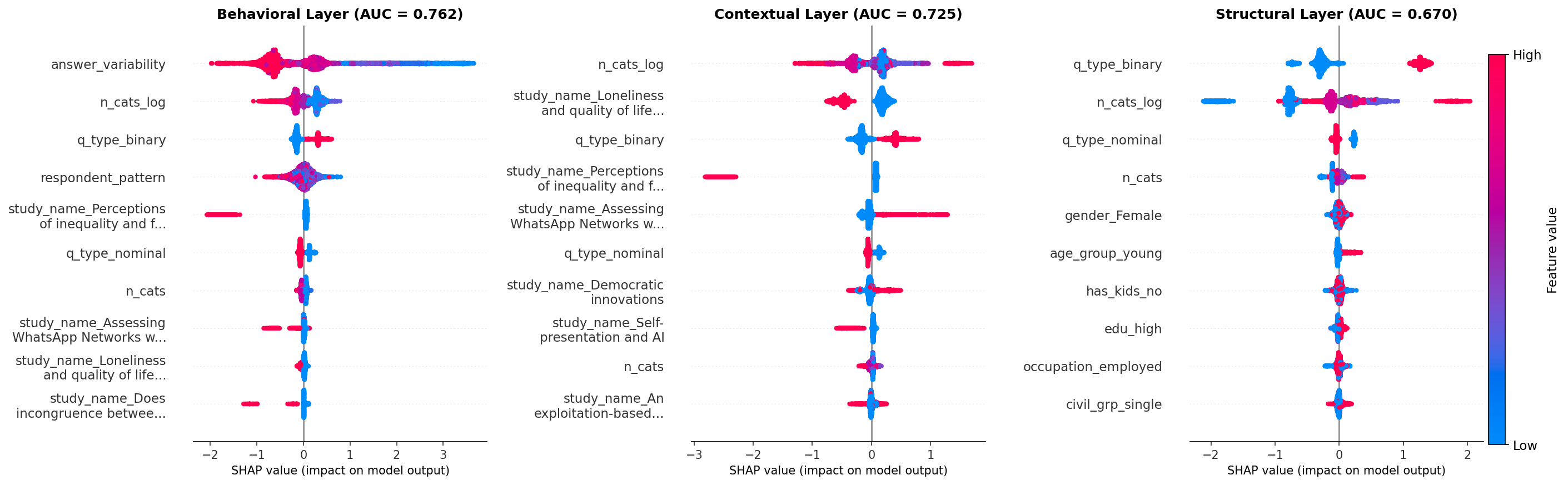}
    \caption{
   \small SHAP beeswarm plots showing the top ten predictors of digital persona accuracy (single-wave prediction) across three feature layers.
    }
    \label{fig:shap_sw}
\end{figure}

\begin{table}[ht]
\centering
\scriptsize
\caption{\small Fidelity of digital personas on single-wave predictions across five persona conditions and three LLMs. Metrics and notation follow Table~\ref{tab:cs_dimensions}. Bold denotes values statistically tied with the best-performing setting under overlap of the reported uncertainty intervals.}
\label{tab:sw_dimensions}
\begin{tabular}{llccccc}
\toprule
\multicolumn{2}{l}{} & \textbf{Baseline} & \textbf{Background} & \textbf{Profile} & \textbf{Profile\texttt{+}Lexical} & \textbf{Profile\texttt{+}Semantic} \\
\midrule
\multirow{3}{*}{\textbf{Clustering ARI} ($\uparrow$)}
& \textbf{GPT} & 0.067 ($\pm$0.014) & 0.044 ($\pm$0.017) & 0.145 ($\pm$0.020) & 0.127 ($\pm$0.026) & 0.169 ($\pm$0.026) \\
& \textbf{Claude} & --- & 0.076 ($\pm$0.029) & \textbf{0.316 ($\pm$0.056)} & 0.216 ($\pm$0.037) & 0.209 ($\pm$0.034) \\
& \textbf{Gemini} & --- & 0.064 ($\pm$0.024) & 0.125 ($\pm$0.020) & 0.129 ($\pm$0.023) & 0.126 ($\pm$0.021) \\
\midrule

\multirow{3}{*}{\textbf{Resp. Match rate} ($\uparrow$)}
& \textbf{GPT} & 0.445 ($\pm$0.003) & 0.446 ($\pm$0.003) & 0.455 ($\pm$0.003) & 0.451 ($\pm$0.003) & 0.452 ($\pm$0.003) \\
& \textbf{Claude} & --- & 0.430 ($\pm$0.003) & 0.442 ($\pm$0.003) & 0.442 ($\pm$0.003) & 0.443 ($\pm$0.003) \\
& \textbf{Gemini} & --- & 0.458 ($\pm$0.003) & \textbf{0.467 ($\pm$0.003)} & 0.459 ($\pm$0.003) & \textbf{0.462 ($\pm$0.003)} \\
\midrule

\multirow{3}{*}{\textbf{Question F1 score} ($\uparrow$)}
& \textbf{GPT} & 0.303 ($\pm$0.004) & 0.358 ($\pm$0.004) & 0.373 ($\pm$0.003) & 0.373 ($\pm$0.004) & 0.374 ($\pm$0.003) \\
& \textbf{Claude} & --- & 0.347 ($\pm$0.004) & 0.359 ($\pm$0.003) & 0.358 ($\pm$0.003) & 0.366 ($\pm$0.003) \\
& \textbf{Gemini} & --- & 0.361 ($\pm$0.005) & \textbf{0.379 ($\pm$0.004)} & \textbf{0.378 ($\pm$0.004)} & \textbf{0.382 ($\pm$0.004)} \\
\midrule

\multirow{3}{*}{\textbf{Ques.~Dist.~JSD} ($\downarrow$)}
& \textbf{GPT} & 0.537 ($\pm$0.003) & 0.416 ($\pm$0.004) & 0.407 ($\pm$0.004) & \textbf{0.394 ($\pm$0.004)} & \textbf{0.399 ($\pm$0.004)} \\
& \textbf{Claude} & --- & 0.435 ($\pm$0.003) & 0.415 ($\pm$0.004) & 0.420 ($\pm$0.004) & \textbf{0.399 ($\pm$0.004)} \\
& \textbf{Gemini} & --- & 0.451 ($\pm$0.004) & 0.417 ($\pm$0.004) & 0.408 ($\pm$0.004) & 0.404 ($\pm$0.004) \\
\midrule

\multirow{3}{*}{\textbf{Resp.~Dist.~MMD} ($\downarrow$)}
& \textbf{GPT} & 0.292 ($\pm$0.003) & 0.235 ($\pm$0.003) & 0.230 ($\pm$0.003) & \textbf{0.216 ($\pm$0.003)} & 0.225 ($\pm$0.003) \\
& \textbf{Claude} & --- & 0.254 ($\pm$0.003) & 0.236 ($\pm$0.002) & 0.231 ($\pm$0.002) & 0.230 ($\pm$0.002) \\
& \textbf{Gemini} & --- & 0.249 ($\pm$0.003) & 0.235 ($\pm$0.003) & 0.225 ($\pm$0.003) & 0.227 ($\pm$0.003) \\
\midrule

\multirow{3}{*}{\textbf{Equity DPI} ($\downarrow$)}
& \textbf{GPT} & \textbf{0.021 ($\pm$0.004)} & \textbf{0.029 ($\pm$0.004)} & \textbf{0.024 ($\pm$0.004)} & \textbf{0.023 ($\pm$0.004)} & \textbf{0.021 ($\pm$0.003)} \\
& \textbf{Claude} & --- & 0.032 ($\pm$0.005) & \textbf{0.026 ($\pm$0.004)} & \textbf{0.022 ($\pm$0.004)} & \textbf{0.027 ($\pm$0.004)} \\
& \textbf{Gemini} & --- & \textbf{0.027 ($\pm$0.004)} & \textbf{0.025 ($\pm$0.004)} & \textbf{0.027 ($\pm$0.005)} & \textbf{0.028 ($\pm$0.004)} \\
\bottomrule
\end{tabular}
\end{table}

\subsection{Supplementary Experiments for Section~\ref{sec:modelling}}\label{app:modelling}

To comprehensively assess digital persona prediction accuracy, we fitted the full modeling pipeline under both prediction directions: single-wave prediction and core prediction. For each direction, models were estimated across three feature layers: \textit{behavioral} (including per-question answer variability and per-respondent answer rarity), \textit{contextual} (retaining study domain but removing behavioral aggregates), and \textit{structural} (question format and demographics only, excluding study domain). Within each layer, five model classes were applied: Logistic Regression with clustered standard errors, Mixed-Effects Logistic Regression (MEGLM) with crossed random intercepts for respondent and question, Decision Tree (max depth = 5), Random Forest ($n = 300$ trees), and XGBoost ($n = 400$ estimators) with SHAP-based feature attribution.

Figure~\ref{fig:shap_sw} shows SHAP beeswarm plots for the XGBoost model under single-wave prediction across all three feature layers. Answer variability and the number of response options dominate the behavioral layer, while question format and demographic variables emerge as the primary drivers in the structural layer, a pattern consistent with the core-prediction results in Figure~\ref{fig:shap_cs}.

Tables~\ref{tab:sw_cs_comparison} and~\ref{tab:cs_sw_comparison} summarise the full results. Across both prediction directions, the behavioral layer performs strongest overall, with XGBoost achieving the highest accuracy. The gap between behavioral and structural layers is more pronounced in the direction of core prediction, where richer feature sets confer a larger marginal gain.

\section{Sampling Design}
\label{app:sampling_design}

We construct separate respondent samples for the two prediction tasks, which we refer to as the Core and Single-Wave samples. In both cases, the goal is to preserve demographic variation while selecting respondents with sufficient prior and target answer coverage. Table~\ref{tab:sample_demographics} is the summary of two sample groups.

\paragraph{Eligible population.}
For each task, a respondent is eligible if they have a non-empty background file, non-empty prior-answer history for the relevant input scope, and at least one evaluable target answer. Let $U$ denote the eligible respondent population for a task, with size $N$. Each respondent $i \in U$ is associated with demographic variables and response-count variables, including the number of available prior answers $n_i^{\mathrm{prior}}$ and target answers $n_i^{\mathrm{target}}$.

\paragraph{Demographic strata.}
We stratify respondents by age group, gender, and household stage. Age is grouped into four categories: $18$--$34$, $35$--$49$, $50$--$64$, and $65+$. Household stage is derived from partner status, children, and living situation, and grouped into family with children, couple without children, single without children, and other or unknown. The primary stratum is the joint demographic cell
\[
h = \text{Age group} \times \text{Gender} \times \text{Household stage}.
\]
Sparse strata with fewer than five eligible respondents are excluded from allocation. Table~\ref{tab:allocation_counts} reports the available and allocated counts for each stratum in both samples.

\paragraph{Proportional allocation.}
For each task, we select $n=500$ respondents. Let $N_h$ be the number of eligible respondents in stratum $h$, with $N=\sum_h N_h$. The target allocation for stratum $h$ is
\[
a_h^* = \frac{N_h}{N}n.
\]
Because $a_h^*$ is generally non-integer, we use largest-remainder proportional allocation. Each stratum first receives $\lfloor a_h^* \rfloor$ respondents, and remaining slots are assigned to strata with the largest fractional remainders. Allocations are capped by the available stratum size, so the final allocation satisfies
\[
\sum_h n_h = n,
\qquad
0 \leq n_h \leq N_h.
\]

\begin{table}[t]
\centering
\caption{\small Demographic composition of the two respondent samples.}
\label{tab:sample_demographics}
\begin{tabular}{llcc}
\toprule
Category & Group & Single-wave & Core \\
\midrule
\multirow{2}{*}{Gender}
    & Female & 266 & 271 \\
    & Male & 234 & 229 \\
\midrule
\multirow{4}{*}{Age group}
    & 18--34 & 19.0\% & 21.2\% \\
    & 35--49 & 20.2\% & 20.4\% \\
    & 50--64 & 28.4\% & 27.4\% \\
    & 65+ & 32.4\% & 31.0\% \\
\midrule
\multirow{3}{*}{Household stage}
    & Couple without children & 39.4\% & 38.0\% \\
    & Family with children & 35.0\% & 36.4\% \\
    & Single without children & 25.6\% & 25.6\% \\
\bottomrule
\end{tabular}
\end{table}

\paragraph{Within-stratum selection.}
Within each stratum, respondents are ranked by answer coverage. For respondent $i$ in stratum $h$, define
\[
x_i = n_i^{\mathrm{prior}},
\qquad
y_i = n_i^{\mathrm{target}}.
\]
We compute within-stratum standardized scores
\[
z_i^{(x)} = \frac{x_i-\bar{x}_h}{s_{x,h}},
\qquad
z_i^{(y)} = \frac{y_i-\bar{y}_h}{s_{y,h}},
\]
where $\bar{x}_h$ and $\bar{y}_h$ are stratum means, and $s_{x,h}$ and $s_{y,h}$ are stratum standard deviations. If a standard deviation is zero or undefined, the corresponding standardized score is set to zero. The main ranking score is
\[
s_i = z_i^{(x)} + z_i^{(y)}.
\]
Ties are broken using the product $x_i y_i$, followed by the target and prior answer counts. Before ranking, respondents within each stratum are randomly shuffled with a fixed seed to make tie-breaking reproducible. The top $n_h$ respondents in each stratum are selected.

\paragraph{Allocation accuracy.}
We evaluate how closely each selected sample preserves the demographic composition of the eligible population. For demographic cell $c$, let $\mathrm{Sample}_c$ be the selected count and
\[
\mathrm{Expected}_c =
\frac{\mathrm{Population}_c}{\sum_j \mathrm{Population}_j} n
\]
be the proportional target count. We summarize allocation error using mean absolute discrepancy,
\[
\mathrm{MAD}
=
\frac{1}{C}\sum_{c=1}^{C}
\left|
\mathrm{Sample}_c-\mathrm{Expected}_c
\right|,
\]
and maximum absolute discrepancy,
\[
\mathrm{MaxD}
=
\max_{1\leq c\leq C}
\left|
\mathrm{Sample}_c-\mathrm{Expected}_c
\right|.
\]
As shown in Table~\ref{tab:allocation_accuracy}, both samples have $\mathrm{MAD}=0.2602$ and $\mathrm{MaxD}=0.5275$. Thus, the selected samples closely preserve the eligible population distribution over age group, gender, and household stage, with an average deviation of less than one respondent per demographic cell.

\begin{table}[!t]
\centering
\small
\caption{\small Allocation counts by demographic stratum for the two prediction tasks.}
\label{tab:allocation_counts}
\begin{tabular}{lllrrrr}
\toprule
Age group & Gender & Household stage
& \multicolumn{2}{c}{Single-Wave}
& \multicolumn{2}{c}{Core} \\
\cmidrule(lr){4-5}\cmidrule(lr){6-7}
& & & Available & Allocated & Available & Allocated \\
\midrule
\multirow{7}{*}{18--34}
& \multirow{3}{*}{Female} & Couple without children & 90 & 12 & 159 & 14 \\
& & Family with children & 220 & 28 & 338 & 29 \\
& & Single without children & 91 & 12 & 146 & 13 \\
\cmidrule(lr){2-7}
& \multirow{3}{*}{Male} & Couple without children & 60 & 8 & 98 & 9 \\
& & Family with children & 159 & 21 & 245 & 21 \\
& & Single without children & 68 & 9 & 104 & 9 \\
\cmidrule(lr){2-7}
& Other & Other or unknown & -- & -- & 5 & 0 \\
\midrule
\multirow{6}{*}{35--49}
& \multirow{3}{*}{Female} & Couple without children & 72 & 9 & 88 & 8 \\
& & Family with children & 305 & 39 & 419 & 37 \\
& & Single without children & 77 & 10 & 128 & 11 \\
\cmidrule(lr){2-7}
& \multirow{3}{*}{Male} & Couple without children & 63 & 8 & 104 & 9 \\
& & Family with children & 230 & 30 & 298 & 26 \\
& & Single without children & 82 & 11 & 113 & 10 \\
\midrule
\multirow{7}{*}{50--64}
& \multirow{3}{*}{Female} & Couple without children & 277 & 36 & 362 & 32 \\
& & Family with children & 248 & 32 & 327 & 28 \\
& & Single without children & 126 & 16 & 169 & 15 \\
\cmidrule(lr){2-7}
& \multirow{3}{*}{Male} & Couple without children & 231 & 30 & 294 & 26 \\
& & Family with children & 224 & 29 & 297 & 26 \\
& & Single without children & 136 & 18 & 174 & 15 \\
\cmidrule(lr){2-7}
& Other & Other or unknown & -- & -- & 1 & 0 \\
\midrule
\multirow{6}{*}{65+}
& \multirow{3}{*}{Female} & Couple without children & 288 & 37 & 495 & 43 \\
& & Family with children & 18 & 2 & 36 & 3 \\
& & Single without children & 227 & 29 & 375 & 33 \\
\cmidrule(lr){2-7}
& \multirow{3}{*}{Male} & Couple without children & 406 & 52 & 642 & 56 \\
& & Family with children & 33 & 4 & 56 & 5 \\
& & Single without children & 136 & 18 & 254 & 22 \\
\bottomrule
\end{tabular}
\end{table}

\begin{table}[!t]
\centering
\caption{Allocation accuracy for the two respondent samples.}
\label{tab:allocation_accuracy}
\begin{tabular}{lcccc}
\toprule
Prediction task &
Eligible population &
Sample size &
MAD &
MaxD \\
\midrule
Core & 4,266 & 500 & 0.2602 & 0.5275 \\
Single-Wave & 5,785 & 500 & 0.2602 & 0.5275 \\
\bottomrule
\end{tabular}
\end{table}

\begin{figure}[t]
    \centering
    \includegraphics[width=\textwidth]{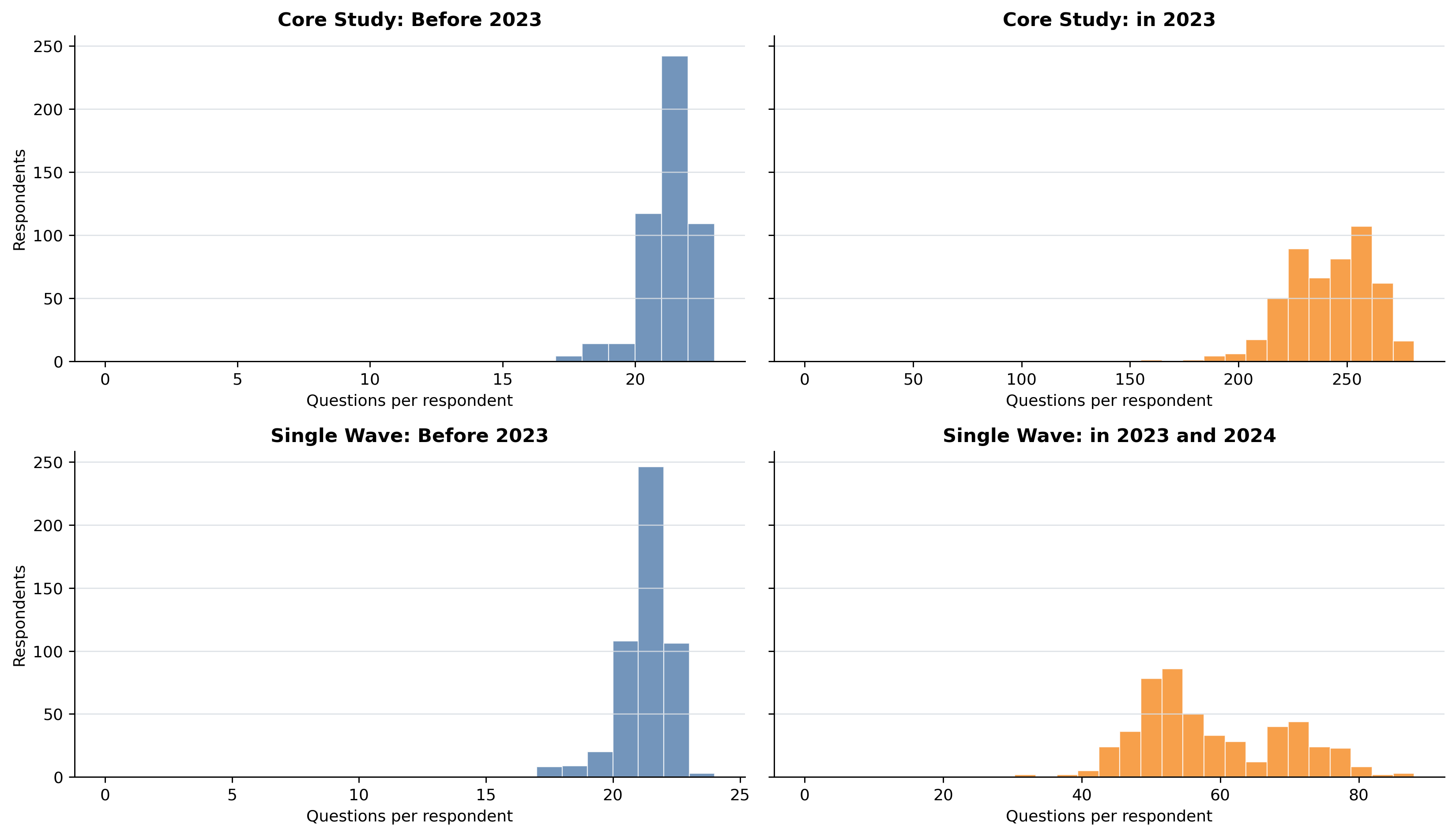}
    \caption{\small
    Distribution of the number of available questions per respondent across the four survey partitions used in the prediction tasks. The left panels show pre-2023 prior-answer coverage for core-study and single-wave histories, while the right panels show held-out target-answer coverage for core-study targets in 2023 and single-wave targets in 2023--2024. The histograms indicate that sampled respondents have dense prior-answer histories and substantial held-out target coverage, supporting respondent-level evaluation.
    }
    \label{fig:question_coverage}
\end{figure}

\paragraph{Final distribution of respondents. }
Figure~\ref{fig:question_coverage} summarizes respondent-level answer coverage in the prior and target partitions used for evaluation after sampling.

\section{Positioning Relative to Prior Work (Expanded)}
\label{app:related-work-positioning}

\paragraph{LLMs as synthetic survey respondents.}
A growing body of work asks whether LLMs can approximate human survey responses and support scalable social measurement. Early studies suggested that LLMs conditioned on demographic profiles can reproduce some population-level response patterns \citep{argyle2023out}. Subsequent work has offered a more cautious picture: synthetic responses may distort variation and statistical relationships even when aggregate averages appear plausible \citep{bisbee2024synthetic}, and LLM survey answers can be sensitive to ordering, labeling, and other design artifacts \citep{dominguez-olmedo2024questioning}. More recent work has explored whether model adaptation can improve aggregate survey simulation, including specialized models for global response distributions and personalized survey-data modeling \citep{cao-etal-2025-specializing,kaiser2025simulating}. Together, these studies show that LLMs can sometimes approximate population-level patterns, but aggregate alignment alone does not establish whether simulated respondents preserve individual answers, subgroup relationships, or scientific conclusions.

\paragraph{Grounded digital personas and personalized user modeling.}
A related literature moves beyond generic demographic prompting toward personas grounded in richer individual information. Prior opinions can improve prediction beyond demographics and ideology alone \citep{hwang-etal-2023-aligning}, and recent digital-persona systems condition models on combinations of demographic, behavioral, psychometric, conversational, or self-report data \citep{chen-etal-2025-personatwin,jiang2025know,park2024generative,toubia2025twin}. These studies show that richer personal histories can improve persona fidelity and make individual-level simulation more plausible. Related meta in-context learning methods learn across groups to align an LLM with group-level preferences, whereas our setting is training-free and evaluates whether personas grounded in individual respondents’ histories preserve held-out survey findings \citep{zhao2024group}. However, prior persona studies leave open which forms of personal history, profile construction, and retrieved memory are useful for predicting held-out responses from the same real individuals, especially in fixed-choice longitudinal survey settings.

\paragraph{Evaluating persona fidelity and scientific reliability.}
Evaluations of digital personas vary widely, including aggregate distribution matching, personalized response selection, behavior-chain prediction, consistency evaluation, persona-effect measurement, validation of LLM simulations, and human-likeness judgments \citep{jiang2025know,li-etal-2025-far,li2025hugagent,abdulhai2026consistently,hu2024persona,hullman2026validating}. These benchmarks are valuable, but scientific use of digital personas requires a stricter test: predictions should be evaluated against real held-out responses from the people the personas are meant to represent. A persona may generate plausible or personalized outputs while still failing to preserve individual answers, question-level reliability, subgroup patterns, or study-level conclusions. This motivates evaluation across target question types, information sources, persona architectures, model families, and survey domains.

\paragraph{Limits, biases, and conditional substitution.}
Recent work also cautions against treating LLM personas as direct substitutes for human participants. LLMs may misportray or flatten identity groups \citep{wang2025large}, introduce systematic downstream biases through persona generation \citep{li2025promise}, or produce heterogeneous effects across items and subgroups \citep{taday2026assessing}. These findings suggest that digital personas should be treated as conditional scientific instruments rather than universal replacements for human data.

Our study follows this view by asking when digital personas grounded in real respondents' prior survey histories can predict held-out responses from the same individuals, and when their failures make real human data indispensable. Table~\ref{tab:related_work_comparison} positions our study against prior work along three dimensions: the data setting, the persona construction strategy, and the evaluation design. We distinguish studies that use objectively scored fixed-choice outcomes, representative stratified samples, same-person temporal holdouts, real longitudinal panels, and prior survey histories; studies that compare persona variants, use retrieval memory, evaluate multiple LLMs, and include a no-context baseline; and studies that assess individual fidelity, aggregate distributions, subgroup equity, joint response structure, and error drivers.

The main pattern is that prior work covers important but partial areas of this design space. Appendix~\ref{app:criteria} defines the comparison criteria, and Appendix~\ref{app:related-work-positioning} provides the full related-work discussion and study-by-study comparison. Several of the closest studies are recent preprints that have not yet appeared in peer-reviewed venues, but the overall pattern is consistent: no single study combines the full set of data, persona-construction, and evaluation elements that we examine here.

\subsection{Comparison criteria}\label{app:criteria}

The comparison in Table~\ref{tab:related_work_comparison} uses fifteen criteria that capture the main methodological and evaluative components of our study.

\begin{description}
    \item[Fixed-choice answers.]
    The target outcomes are constrained to predefined survey response options, such as binary, ordinal, nominal, or multiple-choice categories. This makes evaluation objective and directly comparable across humans and simulated personas, rather than relying only on semantic similarity or subjective plausibility judgments.

    \item[Representative stratified population sample.]
    The study uses a sample designed to represent a broader population, with explicit stratification or population-based sampling. This differs from convenience samples, crowdsourced samples, synthetic-only populations, or small purposive samples.

    \item[Same-person holdout.]
    The study predicts held-out responses for the same individuals whose prior information is used to construct or condition the persona. This is stricter than predicting aggregate distributions or simulating generic members of demographic groups.

    \item[Temporal split.]
    The study separates prior information and target outcomes by time, so that personas are conditioned only on information available before a cutoff and evaluated on later responses. This reduces leakage and better matches real prospective prediction.

    \item[Real longitudinal panel.]
    The data come from real respondents observed repeatedly across waves, studies, or time periods. This allows the study to evaluate whether personas preserve individual continuity, rather than only one-time demographic or attitudinal profiles.

    \item[Prior survey history.]
    The persona is grounded in respondents' previous survey answers, not only in demographics, short descriptions, interviews, or manually written profiles. This criterion captures whether the model receives structured evidence about the person's past choices.

    \item[Persona variants.]
    The study compares more than one way of constructing or prompting personas, such as background-only prompts, generated profiles, profile-plus-memory variants, or other alternative conditioning schemes.

    \item[Retrieval memory.]
    The persona has access to a retrieved subset of relevant prior information at prediction time. This differs from giving a static profile or all background information without question-specific retrieval.

    \item[Multi-LLM comparison.]
    The study evaluates more than one language model family or model variant, allowing results to distinguish persona-method effects from model-specific effects.

    \item[No-context baseline.]
    The study includes a baseline in which the model answers without respondent-specific information. This tests whether persona grounding improves prediction beyond what can be inferred from the question wording and answer options alone.

    \item[Individual fidelity.]
    The study evaluates whether simulated personas reproduce the held-out answers of specific individuals. This includes metrics such as exact accuracy, F1, or respondent-level predictive performance.

    \item[Aggregate distribution.]
    The study evaluates whether simulated responses reproduce population-level or question-level response distributions, even when individual-level prediction may be imperfect.

    \item[Equity / subgroups.]
    The study examines whether performance differs across demographic or social groups, or whether persona simulation introduces, amplifies, or reduces subgroup disparities.

    \item[Joint structure.]
    The study evaluates whether simulated responses preserve multivariate relationships across questions or respondents, such as clustering, correlation structure, latent groups, or joint response patterns.

    \item[Error drivers.]
    The study analyzes which factors explain prediction errors, such as question type, answer rarity, response entropy, domain, respondent characteristics, or model architecture.
\end{description}

Overall, Table~\ref{tab:related_work_comparison} indicates that our contribution is not simply the use of LLMs for survey simulation, nor simply the construction of respondent-level personas. The contribution is the combination of a representative longitudinal panel, prior survey histories, fixed-choice held-out targets, same-person temporal prediction, multiple persona architectures, retrieval-augmented memory, multiple model families, a no-context baseline, and evaluation across individual, aggregate, subgroup, joint-structure, and error-driver dimensions.

\section{Computational Resources and Cost}\label{app:comp_cost}

Running the full experimental pipeline cost approximately \$3{,}000 in API usage, including \$1{,}700 for GPT-5.4, \$600 for Claude Haiku 4.5, and \$700 for Gemini 3 Flash Preview. In addition, we used GPT-5.4 to generate respondent profiles at an average cost of approximately \$0.30 per respondent. The full run took approximately 80 hours on a single 8-core laptop. For each LLM, we evaluated four persona architectures on 500 respondents in each of the two prediction tasks, yielding 4{,}000 respondent-architecture-task predictions per LLM. This corresponds to an average API cost of approximately \$0.4 per GPT-5.4 prediction, \$0.15 per Claude prediction, and \$0.18 per Gemini prediction. Equivalently, running all four persona architectures across both tasks costs approximately \$3.40 per respondent for GPT-5.4, \$1.20 for Claude Haiku 4.5, and \$1.40 for Gemini 3 Flash. By construction, token usage increases from Background to Profile and retrieval-augmented architectures as additional respondent context is included. Retrieval-augmented variants provide only modest performance gains over Profile. For larger populations, retrieval costs scale approximately linearly with the number of respondents, while embeddings are computed once.

\section{Broader Impacts}
\label{app:impact}

This work examines when large language models can reliably substitute for human survey respondents, a question with direct implications for social science research, public opinion measurement, and policy-relevant data collection. By identifying the conditions under which digital personas succeed and fail, our findings help researchers make more informed decisions about when synthetic respondents can appropriately supplement or replace costly human data collection. The clearest benefit is distributional approximation for low-variability questions which suggests targeted applications in instrument pre-testing and early-stage population simulation, where approximate aggregate patterns are sufficient and the cost of fielding a full human sample is prohibitive.

At the same time, this work surfaces risks that deserve attention. Digital personas perform poorly for rare, heterogeneous, and idiosyncratic responses which are precisely the kinds of answers that matter most for studying underrepresented groups, minority opinions, and social heterogeneity. Deploying digital personas without accounting for these failure modes risks systematically flattening the very variation that survey research is designed to capture, potentially misrepresenting minority populations or producing misleading subgroup comparisons. Researchers and practitioners should treat digital personas as conditional instruments requiring task-specific validation rather than as general-purpose replacements for human participants, and should exercise particular caution in applications where individual-level fidelity or subgroup accuracy is scientifically or ethically consequential.

\newpage
\section*{NeurIPS Paper Checklist}

\begin{enumerate}

\item {\bf Claims}
    \item[] Question: Do the main claims made in the abstract and introduction accurately reflect the paper's contributions and scope?
    \item[] Answer: \answerYes{}.
    \item[] Justification: As stated in the question.

\item {\bf Limitations}
    \item[] Question: Does the paper discuss the limitations of the work performed by the authors?
    \item[] Answer: \answerYes{}.
    \item[] Justification: We include the limitations in Section~\ref{sec:conclusion}

\item {\bf Theory assumptions and proofs}
    \item[] Question: For each theoretical result, does the paper provide the full set of assumptions and a complete (and correct) proof?
    \item[] Answer: \answerNA{}.
    \item[] Justification: No theoretical results.

    \item {\bf Experimental result reproducibility}
    \item[] Question: Does the paper fully disclose all the information needed to reproduce the main experimental results of the paper to the extent that it affects the main claims and/or conclusions of the paper (regardless of whether the code and data are provided or not)?
    \item[] Answer: \answerYes{}.
    \item[] Justification: All experimental details and results are provided in Section~\ref{sec:experiments},~\ref{sec:results}--\ref{sec:modelling} and Appendix~\ref{app:experiments_details}--\ref{app:add_exp}.

\item {\bf Open access to data and code}
    \item[] Question: Does the paper provide open access to the data and code, with sufficient instructions to faithfully reproduce the main experimental results, as described in supplemental material?
    \item[] Answer: \answerYes{}.
    \item[] Justification: Details and code are available in Appendix~\ref{app:experiments_details}.

\item {\bf Experimental setting/details}
    \item[] Question: Does the paper specify all the training and test details (e.g., data splits, hyperparameters, how they were chosen, type of optimizer) necessary to understand the results?
    \item[] Answer: \answerYes{}.
    \item[] Justification: See experiment setting and details in Section~\ref{sec:experiments},~\ref{sec:results}--\ref{sec:modelling} and Appendix~\ref{app:experiments_details}--\ref{app:add_exp}..

\item {\bf Experiment statistical significance}
    \item[] Question: Does the paper report error bars suitably and correctly defined or other appropriate information about the statistical significance of the experiments?
    \item[] Answer:\answerYes{}.
    \item[] Justification: We provide specific stratified sampling design with details in Appendix~\ref{app:sampling_design}. We also provide bootstrap confidence intervals for the main results and the modeling in Tables~\ref{tab:cs_dimensions}, \ref{tab:sw_cs_comparison}, and \ref{tab:cs_sw_comparison} (in Appendix).

\item {\bf Experiments compute resources}
    \item[] Question: For each experiment, does the paper provide sufficient information on the computer resources (type of compute workers, memory, time of execution) needed to reproduce the experiments?
    \item[] Answer: \answerYes{}.
    \item[] Justification: We provide a specific section in Appendix~\ref{app:comp_cost}.
    
\item {\bf Code of ethics}
    \item[] Question: Does the research conducted in the paper conform, in every respect, with the NeurIPS Code of Ethics \url{https://neurips.cc/public/EthicsGuidelines}?
    \item[] Answer: \answerYes{}.
    \item[] Justification: As stated in the question.

\item {\bf Broader impacts}
    \item[] Question: Does the paper discuss both potential positive societal impacts and negative societal impacts of the work performed?
    \item[] Answer: \answerYes{}.
    \item[] Justification: This is a subsection in the Appendix~\ref{app:impact}.
    
\item {\bf Safeguards}
    \item[] Question: Does the paper describe safeguards that have been put in place for responsible release of data or models that have a high risk for misuse (e.g., pre-trained language models, image generators, or scraped datasets)?
    \item[] Answer: \answerNA{}.
    \item[] Justification: No risk.

\item {\bf Licenses for existing assets}
    \item[] Question: Are the creators or original owners of assets (e.g., code, data, models), used in the paper, properly credited and are the license and terms of use explicitly mentioned and properly respected?
    \item[] Answer: \answerYes{}.
    \item[] Justification: We provide details on data usage and license in Appendix~\ref{app:liscense}.

\item {\bf New assets}
    \item[] Question: Are new assets introduced in the paper well documented and is the documentation provided alongside the assets?
    \item[] Answer:\answerNA{}.
    \item[] Justification: No new assets.

\item {\bf Crowdsourcing and research with human subjects}
    \item[] Question: For crowdsourcing experiments and research with human subjects, does the paper include the full text of instructions given to participants and screenshots, if applicable, as well as details about compensation (if any)? 
    \item[] Answer: \answerNA{}.
    \item[] Justification: No crowdsourcing nor research with human subjects besides usage of already available data.

\item {\bf Institutional review board (IRB) approvals or equivalent for research with human subjects}
    \item[] Question: Does the paper describe potential risks incurred by study participants, whether such risks were disclosed to the subjects, and whether Institutional Review Board (IRB) approvals (or an equivalent approval/review based on the requirements of your country or institution) were obtained?
    \item[] Answer: \answerNA{}.
    \item[] Justification: No crowdsourcing nor research with human subjects besides usage of already available data.

\item {\bf Declaration of LLM usage}
    \item[] Question: Does the paper describe the usage of LLMs if it is an important, original, or non-standard component of the core methods in this research? Note that if the LLM is used only for writing, editing, or formatting purposes and does \emph{not} impact the core methodology, scientific rigor, or originality of the research, declaration is not required.
    \item[] Answer: \answerYes{}.
    \item[] Justification: Digital personas are fully LLM dependent, this is stated clearly in the manuscript and further detailed in the Appendix~\ref{app:experiments_details}.

\end{enumerate}

\end{document}